%% file: main.tex
\documentclass[10pt,journal,compsoc]{IEEEtran}
\newif\ifbluetext
\bluetextfalse

\newcommand{\mytext}[1]{%
    \ifbluetext
        \textcolor{blue}{#1}%
    \else
        #1%
    \fi
}

\newif\iforangetext
\orangetexttrue

\newcommand{\getext}[1]{%
    \ifbluetext
        \textcolor{blue}{#1}%
    \else
        #1%
    \fi
}

\usepackage{amsmath,amsfonts}
\usepackage{algorithmic}
\usepackage{algorithm}
\usepackage{array}
\usepackage[caption=false,font=normalsize,labelfont=sf,textfont=sf]{subfig}
\usepackage{textcomp}
\usepackage{stfloats}
\usepackage{url}
\usepackage{verbatim}
\usepackage{graphicx}
\usepackage{cite}
\usepackage{booktabs}       
\usepackage{amsfonts}       
\usepackage{nicefrac}       
\usepackage{microtype}      
\usepackage{xcolor}         
\usepackage{graphicx}
\usepackage{multirow}
\usepackage{siunitx}
\usepackage{mathrsfs}
\usepackage{hyperref}
\usepackage{makecell}
\usepackage{upgreek}
\usepackage{colortbl}

\ifCLASSINFOpdf
\else
\fi
\begin{document}
%
\title{Surfel-based Gaussian Inverse Rendering for Fast and Relightable Dynamic Human
Reconstruction from Monocular Videos}
%
%
%
%

\author{Yiqun Zhao, Chenming Wu, Binbin Huang, Yihao Zhi, Chen Zhao,\\ Jingdong Wang,~\IEEEmembership{Fellow,~IEEE}, and Shenghua Gao,~\IEEEmembership{Senior~Member,~IEEE}

\IEEEcompsocitemizethanks{
\IEEEcompsocthanksitem This work is supported by the General Research Fund  of the Research Grants Council (grant \#17200725), the NSFC \#62172279, and in part by the JC STEM Lab of Robotics for Soft Materials funded by The Hong Kong Jockey Club Charities Trust. We also acknowledge the support
provided by the HKU-Shanghai ICRC. (Corresponding author: Shenghua Gao.)
\IEEEcompsocthanksitem Yiqun Zhao, Binbin Huang, and Shenghua Gao are with the University of Hong Kong, Hong Kong SAR, China. E-mail: gaosh@hku.hk 
\IEEEcompsocthanksitem Chenming Wu, Chen Zhao, and Jingdong Wang are with the Department of Computer Vision Technology (VIS), Baidu Inc., Beijing, China.
\IEEEcompsocthanksitem Yihao Zhi is with the School of Science and Engineering, The Chinese University of Hong Kong, Shenzhen, China.
}
\thanks{Manuscript received April 19, 2005; revised August 26, 2015.}}
\markboth{Journal of \LaTeX\ Class Files,~Vol.~14, No.~8, August~2015}%
{Shell \MakeLowercase{\textit{et al.}}: Bare Demo of IEEEtran.cls for Computer Society Journals}
%



\IEEEtitleabstractindextext{%
\begin{abstract}

Efficient and accurate reconstruction of a relightable, dynamic clothed human avatar from a monocular video is crucial for the entertainment industry. This paper presents SGIA (Surfel-based Gaussian Inverse Avatar), which introduces efficient training and rendering for relightable dynamic human reconstruction. SGIA advances previous Gaussian Avatar methods by comprehensively modeling Physically-Based Rendering (PBR) properties for clothed human avatars, allowing for the manipulation of avatars into novel poses under diverse lighting conditions. Specifically, our approach integrates pre-integration and image-based lighting for fast light calculations that surpass the performance of existing implicit-based techniques. To address challenges related to material lighting disentanglement and accurate geometry reconstruction, we propose an innovative occlusion approximation strategy and a progressive training approach. Extensive experiments demonstrate that SGIA not only achieves highly accurate physical properties but also significantly enhances the realistic relighting of dynamic human avatars, providing a substantial speed advantage. We exhibit more results in our project page: \href{https://GS-IA.github.io}{https://GS-IA.github.io}.
\end{abstract}

}

\maketitle

\IEEEdisplaynontitleabstractindextext

%
\IEEEpeerreviewmaketitle

\input{Sec/1_introduction}

\input{Sec/2_related_work}
\input{Sec/3_methods}

\input{Sec/4_experiments}
\input{Sec/6_discussion}
\input{Sec/5_conclusion}

\ifCLASSOPTIONcompsoc
  \section*{Acknowledgments}
\else
  \section*{Acknowledgment}
\fi
The authors would like to thank Umar Iqbal and Shaofei Wang for providing details about preprocessing RANA dataset, Shaofei Wang for providing the scripts of the camera trajectory setting for free viewpoint rendering on CAPE dataset, Zibo Zhao and Shenhan Qian for the valuable discussion and feedbacks at the initial stage of the project, Yahao Shi and Yanmin Wu for providing the implementation details of GIR, Jingwei Xu for providing the guidance of the draft pipeline figure.

\bibliographystyle{IEEEtran}
\bibliography{reference}    
\ifCLASSOPTIONcaptionsoff
  \newpage
\fi



%

%
\vspace{-21pt}
\end{document}

%% file: Sec/1_introduction.tex
\IEEEraisesectionheading{\section{Introduction}}

\begin{figure*}
    \centering
    \includegraphics[width=\textwidth]{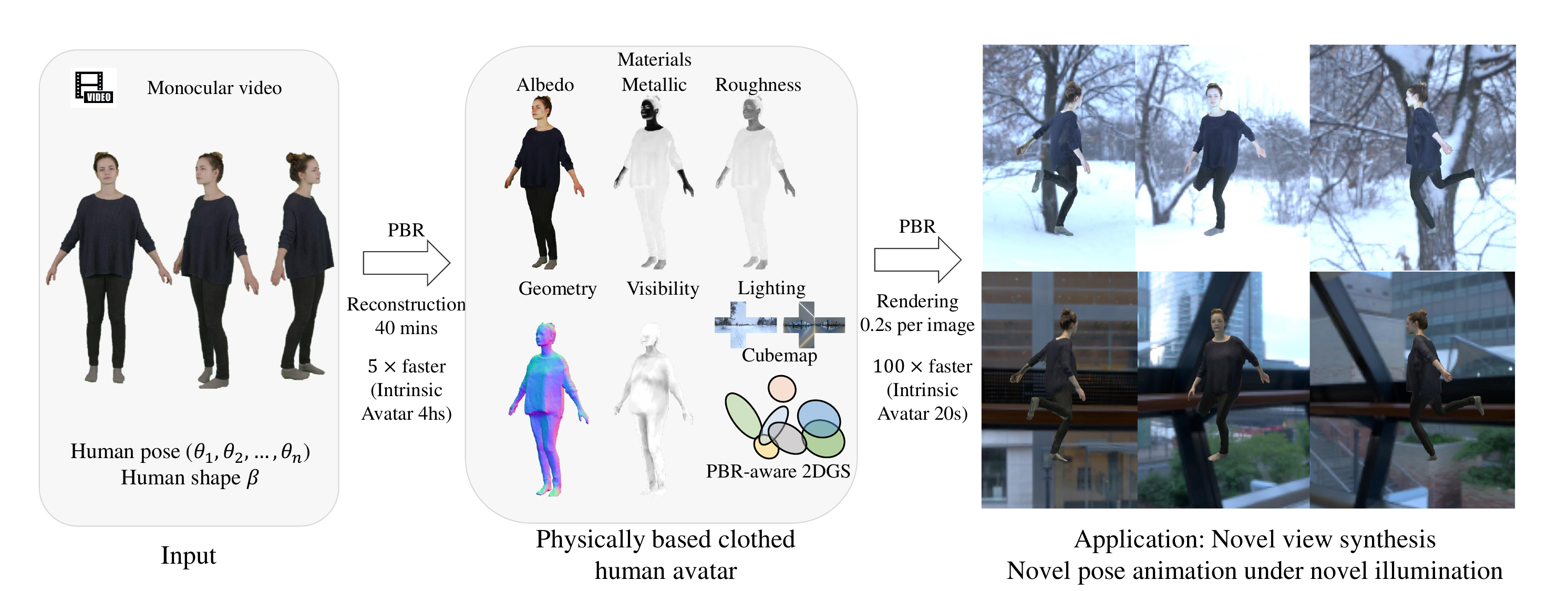}
    \vspace{-8mm}
    \caption{We achieve fast reconstruction of clothed human avatars with PBR properties from a monocular video. SGIA takes a monocular video and initial human pose and shape as input to estimate dynamic clothed humans' PBR properties, including geometry and materials. Leveraging a PBR-aware 2DGS representation, our method enables fast training and rendering processes. By utilizing the estimated PBR properties, we can not only deform the avatars into different poses but also render them with realistic lighting conditions, allowing for versatile and visually appealing outputs.}
    \label{fig:teaser}
    \vspace{-5mm}
\end{figure*}
\IEEEPARstart{R}{econstructing} a clothed human avatar and estimating its Physically-Based Rendering (referred to as PBR) properties simultaneously from monocular videos is a challenging task in computer vision and graphics. It involves disentangling geometry, PBR materials, and environment illumination. With accurate PBR properties, avatars can be deformed into different poses and re-rendered under novel illuminations, as shown in Fig.~\ref{fig:teaser}, offering significant potential for virtual reality, movies, and games.

Early works aim to estimate PBR properties from single images~\cite{Barron2014, Li2018, Li2020, Lichy2021, Sang2020, Wei2020, Yu2019} or multi-view inputs~\cite{Goel2020, Guo2019, Laffont2013, Lensch2003, Park2020, Philip2019, Schmitt2020, Zhang2021}, often assuming known lighting environment maps or scene geometry, which is impractical. Radiance fields~\cite{mildenhall2020nerf} have emerged as effective 3D-consistent scene representations. Works in this direction~\cite{nerv2021, boss2021nerd, knodt2021neural, yang2022psnerf, boss2021neuralpil, physg2021, zhang2021nerfactor, Jin2023TensoIR} estimate geometry, material, and environment light for static scenes from multi-view images but do not account for dynamics, crucial for modeling clothed human avatars. Recent works~\cite{chen2022relighting, lin2024relightable, sun2023neural} have advanced clothed human animation by modeling radiance fields and PBR properties for dynamic humans, but NeRF (Nerural Radiance Fields) optimizes density rather than geometry directly~\cite{wang2021neus}, leading to suboptimal geometry quality. Some works propose neural implicit surfaces~\cite{wang2024intrinsicavatar, xu2024relightable} for better geometry, but all these methods rely on dense point sampling from MLPs~\cite{mildenhall2020nerf} or multi-resolution hash-grids~\cite{mueller2022instant}, slowing down training and rendering due to extensive query operations.

Recently, Gaussian Splatting~\cite{kerbl3Dgaussians, Huang2DGS2024} has revolutionized high-quality, real-time rendering through its explicit representation and efficient splatting-based rasterization on modern GPUs. \mytext{Unlike prior methods~\cite{instant_nvr, jiang2022instantavatar, wang2024intrinsicavatar} that rely on complex backward skinning~\cite{Chen2023PAMI, chen2021snarf} for transformations to a canonical space, clothed human avatars based on Gaussian primitives can be animated easily and efficiently through forward skinning}, akin to the template mesh~\cite{SMPL}. Recent studies~\cite{Lei2024GART, li2024animatablegaussians, zheng2024gpsgaussian, GauHuman, qian20233dgsavatar, yuan2023gavatar, zhu2023ash} have applied Gaussian splatting techniques to reconstruct dynamic clothed human avatars from monocular or multi-view videos, achieving remarkable rendering quality and speed. However, these approaches are limited to capturing radiance fields and fail to recover the full Physically-Based Rendering (PBR) properties of dynamic humans, thus lacking generalization to novel illuminations.

Our goal is to efficiently and accurately reconstruct a \mytext{relightable and dynamic} clothed human avatar from a monocular video, addressing two key challenges: \emph{i.e.}, disentangling coupled materials and shadow effects caused by lighting occlusion, and achieving high-quality geometry of the clothed human avatar. To this end, we propose the Surfel-based Gaussian Inverse Avatar (SGIA), inspired by the efficiency of Gaussian representation. \getext{ SGIA recovers the radiance field and PBR properties of clothed avatars from monocular videos, enabling adaptation to unknown lighting. } We define a set of canonical \mytext{PBR-aware 2DGS (2D Gaussian Splatting)~\cite{Huang2DGS2024}} in canonical space, \getext{transformed to observation space via forward linear blend skinning based on poses}. To accurately model the dynamics of clothed human avatars, we combine articulation from the template model Skinned Multi-Person Linear Model (SMPL)\cite{SMPL} with learnable latent bones conditioned on human poses, inspired by GART~\cite{Lei2024GART}.

To effectively disentangle the coupled materials and shadow effects caused by lighting occlusion, inspired by previous works~\cite{liang2023gs, shi2023gir, jiang2023gaussianshader, R3DG2023}, we add learnable parameters, such as albedo, roughness, and metallic, to each Gaussian for the human avatar. \mytext{ Occlusion modeling can be efficiently handled with spherical harmonics in static scenes~\cite{shi2023gir, liang2023gs}, where the occlusion relationships remain unchanged over time. However, it becomes challenging in dynamic scenes where occlusion relationships change over time. } Directly calculating occlusion with ray tracing~\cite{Bolanos2024GaussianShadowCasting} for each Gaussian is accurate but inefficient. To achieve efficient occlusion calculations, we use modern ray-casting algorithms\mytext{~\cite{libigl-python-bindings-chapter5}} on the limited vertices of the template mesh. \mytext{Our hybrid strategy integrates the advantages of both point and mesh representations: we query the nearest points on the template mesh for each Gaussian and utilize surface points along with normal vectors to bake ambient occlusion from the template mesh. Due to the very small occlusion difference between the clothed human and the unclothed SMPL~\cite{SMPL}, our approach strikes a balance between efficiency and accuracy, effectively disentangling materials from lighting.}

To accurately reconstruct the surface geometry of clothed human avatars, which is crucial for physically based inverse rendering, we propose to leverage the 2DGS~\cite{Huang2DGS2024} as our primitive representation due to its surface-aligned nature, improving geometry reconstruction quality. Each 2D surfel's orientation serves as a local normal direction. Initially, we \mytext{optimize} the rough shape using the original Gaussian color. The geometry remains inaccurate after image reconstruction, especially under high light conditions (Fig.\ref{fig:training_process}). Unlike previous works\cite{liang2023gs, R3DG2023, jiang2023gaussianshader} that freeze Gaussian shape after the first stage, we refine the shape attributes during the PBR optimization stage, leveraging the normal-correlated appearance to enhance geometry. While the original normal consistency loss in 2DGS maintains a splat normal consistency with geometry, it can hinder normal optimization. To address this, we propose a progressive optimization strategy that achieves both accurate geometry and consistent splat normals.

By utilizing the materials, occlusion, and geometry for each splat, we calculate the PBR color of each Gaussian in the observation space and apply the splatting process~\cite{zwicker2004perspective, Huang2DGS2024} designed for 2DGS to produce the PBR image. To address the ambiguity between lighting and materials and achieve plausible PBR material solutions, we use regularization terms to maintain material smoothness on planar surfaces and ensure natural lighting. Extensive experiments on synthetic and real-world datasets show that our method enables fast, relightable clothed human avatar reconstruction within 40 minutes and renders at $540 \times 540$ resolution at 5 FPS, which is $100 \times$ faster than state-of-the-art NeRF-based human inverse rendering methods. Quantitative evaluations demonstrate that our method achieves comparable \mytext{PBR properties} estimation accuracy to state-of-the-art methods. Comprehensive ablation studies further validate our approach.


In summary, our main contributions are:
\begin{itemize}
    \item We propose SGIA, a surfel-based PBR properties modeling method for fast and relightable dynamic clothed human reconstruction from monocular videos.
    \item Our method tackles the challenges in material lighting disentanglement and accurate geometry
    reconstruction by utilizing our proposed occlusion approximation strategy and progressive training strategy.
    \item Extensive experiments demonstrate that our method achieves fast training and rendering while maintaining high-accuracy in PBR properties estimation and human relighting at novel poses.
\end{itemize}

%% file: Sec/2_related_work.tex
\section{Related Work}
\subsection{Inverse rendering for static scenes}
Estimating the physical properties from images is a highly ill-conditioned problem due to its under-constrained nature. Previous works on \mytext{BTF (Bidirectional Texture Function)~\cite{Bi2020, Boss2020, Gardner2003, Ghosh2009, Guarnera2016} and SVBRDF (Spatially Varying Bidirectional Reflectance Distribution Function)~\cite{Haindl2013, Lensch2003, Schmitt2020, Weinmann2015}} estimation rely on specific assumptions of viewpoints, lighting patterns, or capture setups. Other works explored inverse rendering on single image~\cite{Barron2014, Li2018, Li2020, Lichy2021, Sang2020, Wei2020, Yu2019} or multi-view inputs~\cite{Goel2020, Guo2019, Laffont2013, Lensch2003, Park2020, Philip2019, Schmitt2020, Zhang2021}. \mytext{Recent works~\cite{lyu2022neural, lyu2023diffusion} have explored the radiance transfer technique~\cite{lyu2022neural} or utilized data priors from diffusion model~\cite{lyu2023diffusion} to achieve inverse rendering}. These works can only recover high-quality material for each view but fail to achieve realistic and consistent multi-view relighting results under real-world natural illuminations due to the lack of 3D neural fields. 

With the rapid development of Neural Radiance Fields~\cite{mildenhall2020nerf}, lots of works~\cite{zhang2021nerfactor, nerv2021, boss2021nerd, knodt2021neural, boss2021neuralpil, yang2022psnerf, yao2022neilf, boss2022samurai, physg2021, Zhang_2022_CVPR, liu2022neuray, Zhang2023} based on neural field have been proposed to achieve inverse rendering from multi-view images. Specifically, NeRFactor~\cite{zhang2021nerfactor} enabled full estimation of the physical properties (geometry, albedo, materials, and lighting) by incorporating the effect of shadows. Neural-PIL~\cite{boss2021neuralpil} proposed to approximate pre-integration light integration process with a neural network~\cite{sitzmann2019siren} to achieve fast calculation of outgoing radiance. PhySG~\cite{physg2021} proposed to use mixtures of spherical Gaussians to approximate the light transport efficiently. InvRender~\cite{Zhang_2022_CVPR} proposed to  derive the indirect illumination from a pre-trained NeRF to achieve more accurate light disentangling. TensoIR~\cite{Jin2023TensoIR} took advantage of tensor decomposition~\cite{Chen2022ECCV} to improve the quality of reconstruction. Due to inaccurate geometry modeling in NeRFs, some other work~\cite{liu2023nero, mao2023neuspir, iron_2022, yang2023sireir, wang2024nep} explored neural signed distance function (SDF) rendering to achieve more accurate geometry reconstruction for inverse rendering. Among them, NeRO~\cite{liu2023nero} applied the split-sum to constrain the learning of reflective objects. It achieved high-quality geometry reconstruction for the high reflective objects. The Deep Marching Tetrahedra~\cite{shen2021dmtet} provided an efficient and differentiable mesh extraction from neural implicit surface representation. With the benefits of differentiable mesh representation and the differentiable mesh rasterizer, Nvdiffrec~\cite{Munkberg_2022_CVPR} proposed to conduct light integration on the mesh surface to achieve rapid and photo-realistic physically based rendering. NvdiffrecMC~\cite{hasselgren2022nvdiffrecmc} showed that by exploring the combination of mesh-based Mote-Carlo ray-tracing and multi-importance sampling, the decomposition into shape, materials, and lighting can be further improved. 

More recently, Gaussian Splatting based methods~\cite{kerbl3Dgaussians, Yu2024GOF, Huang2DGS2024, Dai2024GaussianSurfels} have achieved high-quality rendering and real-time rendering speed thanks to the forward mapping splatting process. Some works~\cite{liang2023gs, shi2023gir, jiang2023gaussianshader, R3DG2023} proposed to use either the learnable normal vector~\cite{liang2023gs, R3DG2023} or the shortest axis of the Gaussian~\cite{shi2023gir, jiang2023gaussianshader} to achieve geometry representation for inverse rendering. However, these works only focused on the inverse rendering of the static scene, while our work focused on the more challenging inverse rendering of a dynamic clothed human avatar.

 
\subsection{Clothed human avatar modeling}
In addition to the textured mesh, NeRFs~\cite{mildenhall2020nerf} also became a useful representation for photo-realistic clothed avatar reconstruction~\cite{peng2021animatable, peng2021neural, liu2021neuralactor, zhi2022dual, weng_humannerf_2022_cvpr, jiang2022instantavatar, instant_nvr, huang2024tech, huang2022elicit, ARAH_eccv_2022, debsdf2024} from monocular or multi-view videos. \mytext{NeuS2~\cite{wang2023neus2} achieved high quality human reconstruction from multi-view videos without relying on the human template~\cite{SMPL}.} NeuralBody~\cite{peng2021neural} assigned the latent code on the canonical space SMPL \cite{SMPL} vertices and applied the LBS \cite{SMPL} to transform the clothed human avatars to articulation space. For the ray sampling in NeRF~\cite{mildenhall2020nerf}, the world space points are first transformed to articulated SMPL space to query the radiance. They also relied on time-varied latent to fit the pose-dependent deformation. Animatable NeRF~\cite{peng2021animatable} further proposed a pose-dependent deformation network to fit the clothed human conditioned on human. They also utilized the inverse LBS~\cite{peng2021animatable}, which can transform the sample points from SMPL articulated space to the canonical space to query the radiance, thus achieving accurate and fast radiance field modeling for humans. However, there exists a large ambiguity in the inverse LBS calculation. InstantAvatar~\cite{jiang2022instantavatar} proposed to incorporate root finding algorithms~\cite{chen2021snarf, Chen2023PAMI} and multiresolution hash grids to reduce ambiguity in the calculation process. Recent advances in Gaussian Splatting~\cite{kerbl3Dgaussians} assigned the radiance to explicit points that can be easily animated by manipulating the skinning weight from template mesh. The forward splatting process does not need the transformation from articulated space to canonical space. Therefore, both ambiguous inverse LBS and time-consuming root-finding in Snarf\mytext{~\cite{chen2021snarf}} can be avoided. Most works~\cite{GauHuman, Lei2024GART, qian20233dgsavatar, hu2024gaussianavatar, wen2024gomavatar,zhi2025strugauavatar} explored similar techniques to achieve fast clothed human avatar reconstruction from monocular videos. Unlike these works, our work aims to model the PBR properties rather than the radiance only. We also consider the disentanglement of geometry, materials, and environmental lighting to achieve estimation of the full PBR properties.

\subsection{PBR properties reconstruction of clothed human avatars}
Previous works~\cite{kim2024switchlight, pandey2021total, sun2019single, yeh2022learning} explored neural networks to learn the decomposition of materials and lighting from a single portrait image. Recently, \mytext{high} quality physical properties reconstruction of clothed human avatar~\cite{xu2024relightable, saito2024rgca, chen2022relighting, lin2024relightable, wang2024intrinsicavatar, sun2023neural, iqbal2023rana, relightneuralactor2024,qian2022unif, zhi2022dual} have been widely explored with the development of Neural Radiance Fields~\cite{mildenhall2020nerf}. DS-NeRF~\cite{zhi2022dual} proposed to query the lighting at world space to disentangle the light and texture. Relighing4D~\cite{chen2022relighting} is the first to explore the MLPs representation based on NeuralBody~\cite{peng2021neural} to jointly estimate the shape, lighting, and the albedo of dynamic humans from monocular or sparse videos under unknown illumination by disentangling visibility via MLPs. RANA~\cite{iqbal2023rana} pre-trained a mesh representation on ground truth 3d assets to approximate the radiance transfer under different lighting. Relightable Avatar~\cite{lin2024relightable} pretrained an occlusion approximation neural network on a large-scale motion sequence dataset~\cite{ClothCap, AMASS_ICCV_2019}. Intrinsic Avatar \cite{wang2024intrinsicavatar} implemented a fast secondary ray-tracing at articulated space to calculate the visibility and the indirect illuminations. Different from these works, our method explores a relightable clothed human avatar representation based on Gaussian splatting~\cite{kerbl3Dgaussians} to achieve fast training and fast rendering. Concurrent with our work, Animatable Gaussians~\cite{li2024animatable} proposed a relighting extension that predicts occlusion based on the human pose and position map with the 2D StyleUnet~\cite{Karras2019stylegan2} to achieve high-quality rendering. However, their work relied on multi-view videos as input and complex poses during training to achieve pose generalization, which is not suitable for our setting. PhysAvatar~\cite{PhysAvatar2024} also explored the relightable Gaussian Avatar, but it relied on multi-view videos as input to reconstruct the mesh at the first timestamp. 

%% file: Sec/3_methods.tex
\section{Our Method}

\begin{figure*}[htp]
    \centering
    \includegraphics[width=\textwidth]{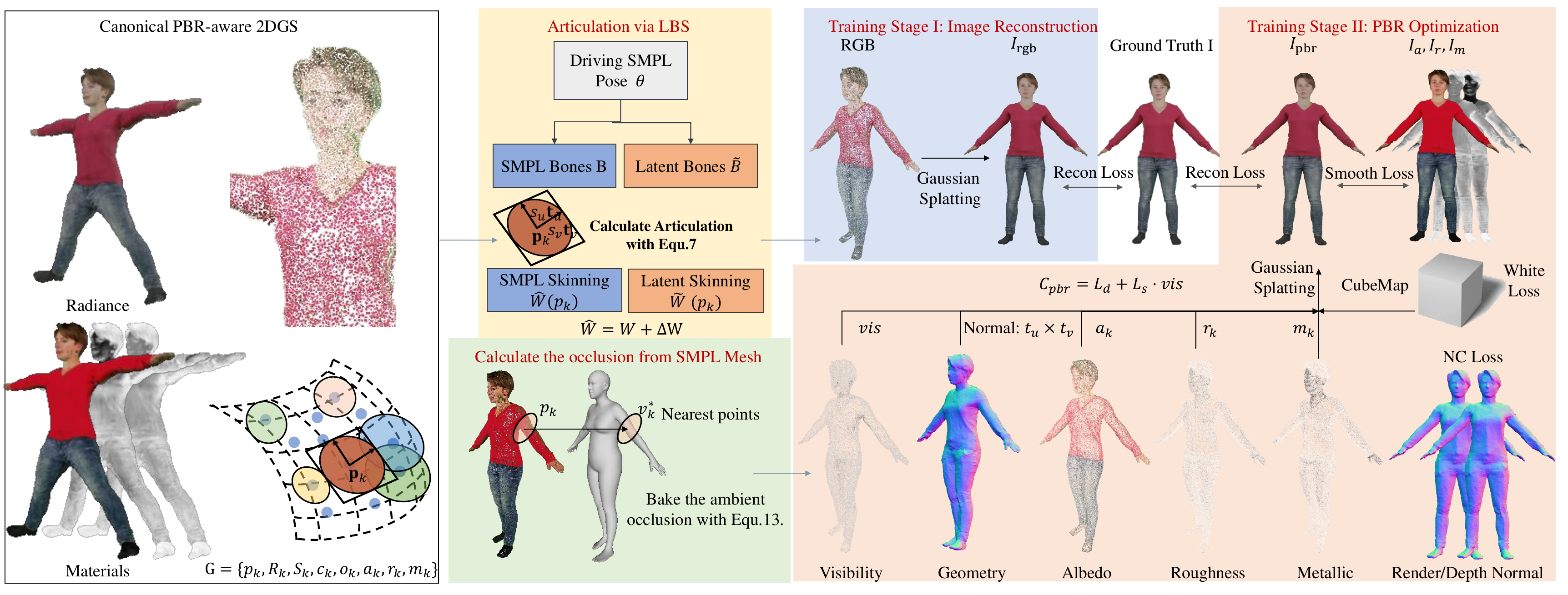}
    \vspace{-8mm}
    \caption{
     Overview of our pipeline. We define the radiance ($c_k$) and materials ($a_k, r_k, m_k$) at the canonical space as canonical PBR-aware 2DGS and deform them to the world space via Linear Blend Skinning (LBS). We first optimize Image reconstruction loss to get the initial shape of the clothed avatar. Based on the rough shape, we optimize the Gaussian attributes with Physically-Based Rendering. To model the shadow effect and decouple the materials from lighting, we propose to approximate the calculation of occlusion with the template mesh. Additionally, we apply regularization, \emph{i.e.}, smooth loss, white loss, and normal consistency loss to get a plausible solution for the PBR materials. 
    }
    \label{fig:pipeline}
    \vspace{-5mm}
\end{figure*}
\mytext{In this section, we introduce our SGIA framework (shown in Fig.\ref{fig:pipeline}) for fast and relightable dynamic clothed human reconstruction from monocular videos. Our method builds upon surfel-based representation\cite{Huang2DGS2024} as described in Sec.~\ref{subsec:revi}.
}
\mytext{
We begin by defining our proposed human representation in Sec.~\ref{subsec:def}. Next, we introduce \textbf{our occlusion calculation strategy}, which enables efficient disentanglement of light and material properties, as detailed in Sec.~\ref{subsec:PBR}. To ensure effective optimization in our framework, we then present \textbf{a novel training strategy} in Sec.~\ref{subsec:opt}. Finally, we describe the animation and relighting process in Sec.~\ref{subsec:infer}.
}

\subsection{A revisit of 2DGS}
\label{subsec:revi}
Unlike 3DGS~\cite{kerbl3Dgaussians}, 2DGS~\cite{Huang2DGS2024} \mytext{adopts} flat 2D Gaussians embedded in 3D space for scene representation. The primitive distributes densities within a planar disk (also known as surfel), which makes the primitive better \mytext{align} with surfaces. This feature can improve the quality of our geometry reconstruction and achieve better material estimation. In 2DGS~\cite{Huang2DGS2024}, each primitive is characterized by \mytext{a central point} $\mathbf{p}_k \in \mathbb{R}^3$ , two principal tangential vectors $\mathbf{t}_u, \mathbf{t}_v \in \mathbb{R}^3$ and two scaling factors $s_u, s_v \in \mathbb{R}$. The scaling factors correspond to the variances of the 2D Gaussian.

A 2D Gaussian is defined in a local tangent plane of the world space, parameterized as :
\begin{equation}
    \mathbf{P}(u, v) = \mathbf{p}_k + s_u\mathbf{t}_uu + s_v\mathbf{t}_vv = [\mathbf{R}_k\mathbf{S}_k , \mathbf{p}_k ] (u, v, 1, 1)^T \text{,}
\end{equation}
\mytext{
where $\mathbf{R}_k = [\mathbf{t}_u, \mathbf{t}_v, \mathbf{t}_u\times \mathbf{t}_v] \in SO(3)$ defines the orientation of each surfel, $\mathbf{t}_u\times \mathbf{t}_v$ represents the surfel's normal vector, and $\mathbf{S}_k = \text{diag}(s_u, s_v, 0)\in \mathbb{R}^{3\times 3}$   is a diagonal matrix encoding the scale. 
}

\mytext{
To render a pixel $x$, 2DGS accumulates the color by first calculating the intersection point between each ray and each tangent plane $\mathbf{P}$, and then accumulating the texture via volumetric alpha blending. The efficient ray-splat intersection calculation process involves finding the intersection between a ray and a local plane in the world space and the algorithm for this is detailed in the original paper~\cite{Huang2DGS2024}. 
}

\mytext{
For each intersection point $x$, we can transform it into the local coordinate defined by the tangent plane $\mathbf{P}$ as $\mathbf{u} = (u, v)$, and calculate its 2D Gaussian value using the following equation:
\begin{equation}
\mathcal{G}(\mathbf{u}) = \exp(-\frac{u^2 + v^2}{2}) \text{,}
\end{equation}
Then, the 2D Gaussians are sorted based on the depth of their centers and organized into tiles.
}

\mytext{
Finally, the color $\mathbf{c}(x)$ at the pixel $x$ can be calculated using the following equation:
\begin{equation}
\mathbf{c}(x) = \sum_{i=1}^n \mathbf{c}_i o_i \mathcal{G}i(\mathbf{u}(x)) \prod_{j=1}^{i-1} (1 - o_j \mathcal{G}_j(\mathbf{u}(x))) \text{,}
\label{eq:splatt}
\end{equation}
where $o_i$ denotes the opacity and $\mathbf{c}_i$ denotes the view-dependent appearance associated with the $i$-th 2D Gaussian.
}


\subsection{Clothed humans avatars as dynamic surfels animated by template models}
\label{subsec:def}
\getext{We build upon recent advancements in modeling humans as animatable Gaussians~\cite{Lei2024GART,hu2023gauhuman, qian20233dgsavatar, li2024animatablegaussians} } Specifically, we define the PBR-aware 2DGS in a canonical space \getext{, and transform them into world space using LBS}. The Gaussians are animated by a human template model~\cite{SMPL} to achieve avatar animation. We choose \getext{a surfel-based primitive~\cite{Huang2DGS2024} as our representation, as it provides improved geometry accuracy.} Furthermore, we define additional parameters for intrinsic properties and lighting on each 2D Gaussian, \getext{ enabling physically based rendering capabilities}.

\noindent\textbf{Canonical PBR-aware 2DGS Representations.} We first utilize a set of PBR-aware 2D Gaussians ($\mathbf{G}$) with material parameters to represent the clothed avatar in a canonical space: 
\begin{equation}
    \mathbf{G} = \{\mathbf{p}_k, \mathbf{R}_k, \mathbf{S}_k, \mathbf{c}_k, o_k, \mathbf{a}_k, r_k, m_k\}_{k=1}^{N_{\text{gs}}} \text{,}
\end{equation}
\getext{Here, $\mathbf{p}_k, \mathbf{R}_k, \mathbf{S}_k, o_k$ are the parameters that defined the shape of each Gaussian and $N_{\text{gs}}$ is the number of the gaussians}. In the first stage, \getext{we optimize the shape parameters of the Gaussians using the appearance information encoded in $\mathbf{c}_k$.} \getext{Then, to achieve physically-based materials estimation and rendering, we introduce additional parameters:} albedo $\mathbf{a}_k \in \mathbb{R}^3$, roughness $r_k \in \mathbb{R}$ and metallics $m_k \in \mathbb{R}$. \getext{These parameters model the decoupled appearance properties on each 2D Gaussian.} \getext{We initialize the location and rotation of our canonical Gaussians using $N_{\text{init}}=6980$ vertices and their corresponding normal vectors from the SMPL template model~\cite{SMPL}, The scaling factors are calculated in the canonical space following the approach proposed in~\cite{Lei2024GART}.}


\noindent\textbf{Articulation via LBS.} We utilize \getext{the SMPL model~\cite{SMPL}  as our articulation template}. The SMPL model is defined by a function $M(\boldsymbol{\theta}, \boldsymbol{\beta}): \mathbb{R}^{|\boldsymbol{\theta}|\times|\boldsymbol{\beta}|}\rightarrow \mathbb{R}^{3N_{\text{init}}}$, where the $\boldsymbol{\theta} \in \mathbb{R}^{3\times K}$  denotes the body pose parameters. 
$\boldsymbol{\beta} \in \mathbb{R}^{10}$ denotes the coefficients for \getext{the body shape}.  

\getext{With the SMPL template model, we can deform the Gaussian primitives to the articulated space using LBS. For each Gaussian point $\mathbf{p}_k$ in the canonical space}, we first calculate the articulation transformation $\mathbf{A}_k = [\mathbf{A}_k^{rot}, \mathbf{A}_k^t]$ \getext{, where $\mathbf{A}_k^{rot}$ denotes the rotation matrix and $\mathbf{A}_k^t$ denotes the translation vector. These  articulation parameters are derived from the pose parameters $\boldsymbol{\theta}$ of the  SMPL model. The transformation $\mathbf{A}_k$ for each Gaussian can be calculated using the following equation:} 
\begin{equation}
    \mathbf{A}_k = \sum_{i=1}^{N_b} \widehat{\mathbf{W}}_i(\mathbf{p}_k) \mathbf{B}_i  \text{,}
    \label{eq:artic}
\end{equation}
$\mathcal{B}(\boldsymbol{\theta}) = [\mathbf{B}_1, \mathbf{B}_2, \dots, \mathbf{B}_{N_b}]$ ($\mathbf{B}_i \in SE(3)$) denotes the rigid transformation that moves each canonical joint to the articulated space. $N_b$ is the number of joints in a template model. $\widehat{\mathbf{W}}(\mathbf{p}_k)$ denotes the blending weight for the bones $\mathcal{B}(\boldsymbol{\theta})$. We model with $\widehat{\mathbf{W}}(\mathbf{p}_k) = \mathbf{W}(\mathbf{p}_k) + \Delta \mathbf{W}(\mathbf{p}_k)$.  $\mathbf{W}(\mathbf{p}_k) \in R^{N_b}$  is the predefined skinning weight queried in canonical space for each Gaussian, and the $\Delta \mathbf{W}(\mathbf{p}_k)$ is the learnable blending weight offset for each Gaussian. 
\getext{Building upon the work of~\cite{Lei2024GART}, we add an additional term to the predefined blending weights $\mathbf{W}$  to model the actual instance deformation.}
Then, we can deform the 2D Gaussians to the articulated space via:
\begin{equation}
    \mathbf{p}_k^* = \mathbf{A}_k^{rot} \mathbf{p}_k + \mathbf{A}_k^t, \quad \mathbf{R}_k^* = \mathbf{A}_k^{rot}\mathbf{R}_k \text{,}
\label{eq:app_rot}
\end{equation}
where $\mathbf{p}_k^*$ and $\mathbf{R}_k^*$  denotes the location and rotation of a 2D Gaussian in articulated space. 

\getext{To approximate the clothing motion captured from the monocular video, we follow the approach proposed in}~\cite{Lei2024GART} and introduce $N_{\text{lb}} = 4$ latent Bones $\widetilde{\mathcal{B}}(\boldsymbol{\theta})$ \getext{parameterized using a MLP conditioned on the human pose} $\boldsymbol{\theta}$. \getext{Additionally, we parameterize the blending weight}
$\widetilde{\mathbf{W}}_i$ for each latent
bones $\widetilde{\mathcal{B}}_k(\boldsymbol{\theta})$. 
\getext{This allows us to extend the articulation transformation equation.~\ref{eq:artic} to the overall articulation:}
\begin{equation}
    \mathbf{A}_k = \sum_{i=1}^{N_b} \widehat{\mathbf{W}}_i(\mathbf{p}_k) \mathbf{B}_i  + \sum_{i=1}^{N_{lb}} \widetilde{\mathbf{W}}_i(\mathbf{p}_k) \widetilde{\mathbf{B}}_i \text{,}
\label{eq:lbs}
\end{equation}

\subsection{Physically-Based Rendering with image-based Lighting}
\label{subsec:PBR}
\noindent\textbf{Physically-Based Inverse Rendering.} Unlike previous physically-based cloth avatars~\cite{wang2024intrinsicavatar, xu2024relightable}, our lighting calculation does not need the time-consuming Monte Carlo Sampling, the integration of ingoing radiance in the hemisphere. We instead utilize the image-based lighting and split-sum approximation~\cite{Karis2013RealShading} to tackle intractable integral.

\getext{The rendering equation calculates the outgoing light from a surfel at the observation space position} $\mathbf{p}_k^*$ \footnote{
\getext{For simplicity, we will omit the subscript $k$ in the following equations, but note that all the terms -  $\mathbf{L}_o$, $\mathbf{L}_i$, $\mathbf{n}$, $\mathbf{a}$, and $m$ correspond to the parameters associated with the specific surfel $\mathbf{p}^*_k$.}
}
\getext{The rendering equation can be written as:}
\begin{equation}
    \mathbf{L}_o(\mathbf{w}_o) = \int_{\Omega} f(\mathbf{w}_o, \mathbf{w}_i) \mathbf{L}_i(\mathbf{w}_i)(\mathbf{w}_i\cdot \mathbf{n}) d\mathbf{w}_i \text{,}
\label{eq:light_o}
\end{equation}
where $\mathbf{w}_o$ denotes the view direction of outgoing radiance $\mathbf{L}_o(\mathbf{w}_o)$, $\mathbf{w}_i$ denotes the direction of the ingoing radiance $\mathbf{L}_i(\mathbf{w}_i)$ sampled from the hemisphere $\Omega$, and $\mathbf{n}$ denotes the normal orientation of the surfel. The BRDFs property $f$ can be divided into  a specular term (the 1st term in equation.~\ref{eq:decomp}) and a diffuse term (the 2nd term in equation.~\ref{eq:decomp}) following the Disney-BRDF model~\cite{Burley2012PhysicallyBasedShading}:
\begin{equation}
    f(\mathbf{w}_o, \mathbf{w}_i) =  \frac{DFG}{4(\mathbf{w}_i\cdot \mathbf{n})(\mathbf{w}_o \cdot \mathbf{n})} + (1-m) \frac{\mathbf{a}}{\pi} \text{,}
    \label{eq:decomp}
\end{equation}
where $D$ denotes the GGX~\cite{GGX} normal distribution function (NDF), $F$ denotes \getext{the Fresnel term~\cite{GGX}}, and $G$ denotes the geometric attenuation,  $\mathbf{a}$ is the albedo for each Gaussian, $m$ denotes the metallics for each Gaussian.
\getext{We further optimize the rendering equation by separating the outgoing radiance $\mathbf{L}_o(\mathbf{w}_o)$ into two distinct parts} (specular light $\mathbf{L}_s$ in equation.~\ref{eq:split-sum} and diffuse light $\mathbf{L}_d$ in equation.~\ref{eq:diffuse}). \getext{We utilize the split-sum approximation proposed in ~\cite{Karis2013RealShading}.  This allows us to separate the integral of the light and BRDF product into two separate integrals, which can then be pre-integrated for efficient querying during rendering.}


For the specular light term $\mathbf{L}_s$, we have
\begin{equation}
    \mathbf{L}_s  \approx \int_{\Omega} \frac{DFG}{4(\mathbf{w}_i\cdot \mathbf{n})(\mathbf{w}_o \cdot \mathbf{n})} d\mathbf{w}_i \int_{\Omega} \mathbf{L}_i(\mathbf{w}_i)D(\mathbf{w}_i, \mathbf{w}_o) (\mathbf{w}_i\cdot \mathbf{n}) d \mathbf{w}_i \text{.}
    \label{eq:split-sum}
\end{equation}
\getext{The first term in the specular light equation represents the integral of the specular BRDF, which depends only on the dot product $\mathbf{w}_i \cdot \mathbf{n}$ and the roughness $r$, independent of the illumination. The second term corresponds to the integral of the incoming radiance and the specular normal distribution function (NDF) $D$. It depends on the reflection direction $\mathbf{r} = 2(\mathbf{w}_i \cdot \mathbf{n})\mathbf{n} - \mathbf{w}_i$ and the roughess $r$, which can be pre-integrated and queried from a filtered cubemap.}

For the diffuse light term $\mathbf{L}_d$, we have: 
\begin{equation}
    \mathbf{L}_d \approx \mathbf{a} (1-m) \int_{\Omega} L(\mathbf{w}_i) \frac{\mathbf{w}_i\cdot \mathbf{n}}{\pi} d\mathbf{w}_i \text{.}
\label{eq:diffuse}
\end{equation}
The integration term can also be efficiently calculated with pre-integration and queried from a cubemap. 

\noindent\textbf{Occlusion Calculation.}  As there exists a lot of self-occlusion during the animation of clothed human avatars, to achieve more realistic physically based rendering, we should consider the occlusion. However, calculating the occlusion with ray-tracing~\cite{Bolanos2024GaussianShadowCasting} for each Gaussian is time-consuming and hinders rendering speed. To achieve occlusion calculation while also maintaining the fast rendering nature of Gaussian Splatting, we proposed to approximate the occlusion as the ambient occlusion from a SMPL~\cite{SMPL} mesh $M(\boldsymbol{\theta}, \boldsymbol{\beta}) = (\mathbf{V}, \mathbf{F})$.  Baking the ambient occlusion of the SMPL mesh can reduce the time for occlusion calculation. Specifically, we first query the nearest surface points $\mathbf{v}_k^*$ from the mesh $M(\boldsymbol{\theta}, \boldsymbol{\beta})$ for each 2D Gaussian $\mathbf{p}_k^*$ at articulated space.  We then utilize the surface points $\mathbf{v}_k^*$ and its corresponding surface normal $\mathbf{n}_k^*$ from the mesh to bake the ambient occlusion for the $k$-th Gaussian:
\begin{equation}
    AO_k = \frac{1}{\pi} \int V_{\mathbf{v}_k^*, \mathbf{w}}(\mathbf{n}_{k}^*\cdot \mathbf{w})d\mathbf{w} \text{,}
\end{equation}
where the $V_{\mathbf{v}_k^*, \mathbf{w}} \in [0, 1] $ is the visibility function at each surface points $\mathbf{v}_k^*$. The integral is approximated by casting rays in $100$ random directions around each vertex. The visibility of the specular light is, therefore, $1-AO_k$. 

The overall outgoing radiance for each Gaussian is:
\begin{equation}
    \mathbf{c}_k^\text{PBR} = \mathbf{L}_o(\mathbf{w}_o) = \mathbf{L}_d + (1-AO_k)\mathbf{L}_s \text{,}
\label{eq:pbr}
\end{equation}
where the $\mathbf{w}_o$ denotes the outgoing direction, corresponding to the view direction for each 2D gaussian in world space. 

\noindent\textbf{Environment Map Representation.} \getext{ We use the $6 \times 64 \times 64$ tensor of trainable parameters to represent the image-based lighting. We query the lighting information at different mipmap levels. The base level corresponds to the pre-integrated lighting for the minimum supported roughness value. The subsequent smaller mipmap levels are derived from this base level using a pre-filtering approach, as described in ~\cite{Karis2013RealShading}.}
\begin{figure}[htp]
    \centering
    \includegraphics[width=0.48\textwidth]{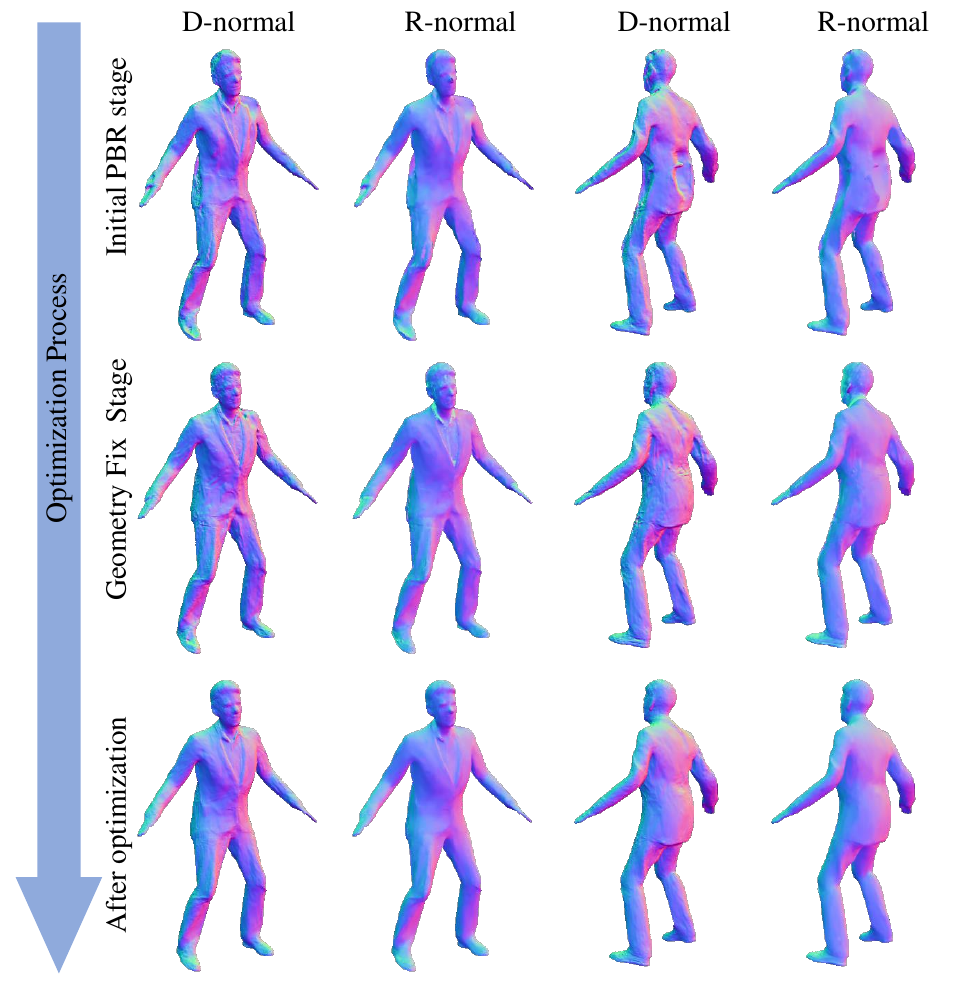}
    \caption{
    Visualization of our PBR optimization stage. At the initial stage, both the D-normal $\mathbf{N}$ (Normal from depth points) and the R-normal (Rendered Splat Normal) are not accurate. After the Geometry Fix Stage ($\lambda_{\text{NC}_2} = 0, \lambda_{\text{NC}_1}=1$). The Geometry is repaired while the splat normal is still messy. We then set ( $\lambda_{\text{NC}_1} = 0, \lambda_{\text{NC}_2}=1$), and finally achieve that both the geometry and splat normal are consistent and accurate.
     }
    \label{fig:training_process}
\end{figure}

\noindent\textbf{Physical-Based Rendering with splatting.} \getext{To render the final color, we first calculate the ray-splat intersections for each pixel and Gaussian primitive. We then apply the following accumulation equation:}
\begin{equation}
    \mathbf{c}^{\text{PBR}}(x) = \sum_{i=1}\mathbf{c}_i^{\text{PBR}}o_i\mathcal{G}_i(\mathbf{u}(x)) \prod_{j=1}^{i-1}(1-o_j\mathcal{G}_j(\mathbf{u}(x))) \text{.}
\label{eq:splatting}
\end{equation}
we can also get the corresponding albedo, roughness, metallic, and visibility maps following the same splatting process of 2DGS.

\subsection{Reconstruct animatable PBR-aware 2DGS from monocular videos}
\label{subsec:opt}

\getext{Our training process consists of two stages. In the first stage, we apply an Image Reconstruction loss to capture the rough shape and appearance of the dynamic humans. This allows us to obtain an initial estimate of the avatar's geometry and appearance. In the second stage, we use PBR techniques to further refine the avatar's material properties and achieve accurate surface reconstruction. Specifically, we optimize the model parameters $\mathbf{G}, \widehat{\mathbf{W}}, \widetilde{\mathbf{B}}, \widetilde{\mathbf{W}}$ using the ground-truth images ${\mathbf{I}_1, ..., \mathbf{I}_M}$ and the corresponding human poses ${\boldsymbol{\theta}_1, ..., \boldsymbol{\theta}_M}$}

\noindent\textbf{Training Stage I -- Image Reconstruction.} In the first stage, we aim to reconstruct the rough shape for clothed avatars with view-dependent color represented as spherical harmonics. We apply the image reconstruction loss and a regularization term on these Gaussian parameters, and the normal-depth consistency loss proposed by~\cite{Huang2DGS2024}. Denote the rendered image as $\hat{\mathbf{I}}$, the training loss at this stage is:
\begin{equation}
    L_{\text{RF}} = \ell_1(\hat{\mathbf{I}}, \mathbf{I}) + \lambda_{\text{SSIM}} L_{\text{SSIM}}(\hat{\mathbf{I}}, \mathbf{I}) + L_{\text{reg}} + \lambda_{\text{NC}} L_{\text{NC}} \text{.}
\label{eq:1stage}
\end{equation}
\getext{We only render the foreground avatar and apply the $L_{\text{RF}}$ loss on it, without modeling the background. Due to the discrete Gaussian representation, the reconstructed cloth avatars may overfit to the sparse 2D observations without regularization. Following prior work ~\cite{Lei2024GART}, we regularize the standard deviation $\sigma_k$ of the Gaussian attributes within the KNN neighborhood of each Gaussian $\mathbf{G}_k$. This leads to:}
\begin{equation}
    L^{\sigma}_k = \sum_{\text{attr}_k \in \{ \mathbf{R}_k, \mathbf{S}_k, o_k, \mathbf{c}_k, \Delta \mathbf{W}_k\}} \lambda_{\text{attr}_k} \sigma(\text{attr}_k) \text{,}
\end{equation}
where $\sigma_{\text{attr}_k}$ denotes the standrd variation of the Gaussian attributes in the KNN neighborhood. Furthermore, we encourage learnable blending weights, and the latent bones for cloth animation for each Gaussian to be small, thus alleviating the overfitting for the cloth deformation. We follow the previous work~\cite{jiang2022instantavatar} to store the learnable blending weights in a 3D voxel and apply a tri-linear interpolation to query the blending weight $\Delta W_k$ for each Gaussian. We therefore regularize the learnable blending weights to be small for both the SMPL bones and latent bones with:
\begin{equation}
    L_k^{\text{norm}} = \lambda_{\widehat{\mathbf{W}}}\|\Delta \mathbf{W}_k\|_2 + \lambda_{\widetilde{\mathbf{W}}} \|\widetilde{\mathbf{W}}_k \|_2 \text{.}
\end{equation}
The total regularization loss is:
\begin{equation}
    L_{\text{reg}} = \frac{1}{N_{\text{gs}}}\sum_{k=1}^{N_{\text{gs}}} L_k^{\sigma} + L_k^{\text{norm}} \text{.}
\end{equation}
To ensure the orientation of each 2D Gaussian  align with the actual surface normal and suitable for physically based rendering. We follow 2DGS~\cite{Huang2DGS2024} and align the splat's normal with the gradients of the depth maps as follows:
\begin{equation}
    L_{\text{NC}} = \sum_i w_i(1-\mathbf{n}_i^T \mathbf{N}) \text{,}
    \label{eq:lnc}
\end{equation}
where $w_i=o_i\mathcal{G}_i(\mathbf{u}(x)) \prod_{j=1}^{i-1}(1-o_j\mathcal{G}_j(\mathbf{u}(x)))$ denotes the blending weight of the $i$-th intersection, $\mathbf{n}_i$ is the normal of the splat facing the camera, $\mathbf{N}$ is normal derived from depth map, \getext{Compared with the $\ell_1$ loss utilized in prior works}~\cite{liang2023gs, jiang2023gaussianshader}, \getext{our normal-consistency loss yields smoother surfaces with more consistent Gaussian orientation along the ray, as shown in our ablation study.}

\noindent\textbf{Training Stage II -- PBR optimization.} At the PBR stage, we first apply the image reconstruction loss on the PBR image. Denote the rendered image of PBR as $\mathbf{\hat{I}}_{\text{pbr}}(\mathbf{a}, m, r, \mathbf{n})$, which is computed from albedo, materials, and splat normal. The overall loss at the PBR stage is:
\begin{equation}
\begin{aligned}
     L_{\text{PBR}} = \ell_1(\hat{\mathbf{I}}_{\text{pbr}}, \mathbf{I}) + \lambda_{\mathbf{a}}L_{\text{smooth}}(\mathbf{a}) +\\
     \lambda_{r} L_{\text{smooth}}(r) + \lambda_{m} L_{\text{smooth}}(m)+ \\
    \lambda_{\text{NC}} L_{\text{NC}^*} + \lambda_{\text{white}} L_{\text{white}}\text{,}
\end{aligned}
\label{eq:2stage}
\end{equation}
where the $L_{\text{NC}^*}$ is slightly different from the $L_{\text{NC}}$ used in the previous stage. With the original normal-consistency loss from the previous stage, we can not reconstruct totally accurate geometry, especially under the extreme high light conditions as shown in Fig~\ref{fig:training_process} and the original 2DGS~\cite{Huang2DGS2024} paper. 
At the PBR stage, the Gaussian appearance \getext{correlates with the splat normal}. \getext{Thus, we can improve geometry during optimization. Different with GS-IR~\cite{liang2023gs}, we cannot freeze Gaussian shapes. However, directly optimizing Eq.~\ref{eq:lnc} leads to inconsistent splat normals due to inaccurate geometry. Discarding the normal consistency regularization makes splat normals erratic Fig.~\ref{fig:ablation_consistency}. Therefore, we propose a progressive optimization strategy for accurate geometry and consistent splat normals:}
\begin{equation}
\begin{aligned}
    L_{\text{NC}^*} = \lambda_{\text{NC}_1}\sum_i w_i(1- \text{sg}(\mathbf{n}_i^T) \mathbf{N})  
    \\+ \lambda_{\text{NC}_2} \sum_i w_i(1- \mathbf{n}_i^T \mathbf{N}) \text{,}
\end{aligned}
\label{eq:progressive}
\end{equation}
where $\text{sg}(\cdot)$ denotes stop gradient. At the initial stage, we set $\lambda_{\text{NC}_2} = 0, \text{and}\, \lambda_{\text{NC}_1}=1$, which means we utilize the splat normal to optimize the geometry $\mathbf{N}$. Then we set $\lambda_{\text{NC}_1} =0, \text{and}\,\lambda_{\text{NC}_2} = 1$ to force the splat normal to be aligned with the fixed geometry. We visualize the splat normal and depth derived normal during the optimization stage at Fig~\ref{fig:training_process}. 
To overcome the ambiguity between the lighting and PBR materials and obtain a plausible solution for PBR materials, we follow ~\cite{shi2023gir} to utilize a smoothness loss which forces the materials and albedo to be smooth at the planar surfaces, where the image gradients are absent. This can be written as:
\begin{equation}
\begin{aligned}
L_{\text{smooth}}(\mathbf{a}) = \frac{1}{N_{\text{pixel}}} \sum_{i,j} |\partial_{\mathbf{x}} \mathbf{a}_{ij}| e^{-\|\partial_{\mathbf{x}} \mathbf{I}_{ij}\|} \\+ \frac{1}{N_{\text{pixel}}} \sum_{i,j} |\partial_{\mathbf{y}} \mathbf{a}_{ij}| e^{-\|\partial_{\mathbf{y}} \mathbf{I}_{ij}\|} \text{,}
\end{aligned}
\label{eq:smooth}
\end{equation}
where $N_{\text{pixel}}$ denotes the numbe of pixels of the image. I represents ground-truth images, $\partial_{\mathbf{x}}, \partial_{\mathbf{y}}$ represents the gradients of an image or material map in $\mathbf{x}$ and $\mathbf{y}$ directions. We apply the smoothness loss on the rendered albedo, metallic and roughness map. Furthermore, we assume a near-natural white incident light to obtain a plausible disentanglement of lighting and materials. We therefore apply the white light regularization on the environment map as:
\begin{equation}
    L_{\text{white}} = \sum_{\text{channel}} (\mathbf{L}_{\text{channel}} - \frac{1}{3} \sum_{\text{channel}} \mathbf{L}_{\text{channel}}), {\text{channel}} \in \{R, G, B\} \text{.}
\label{eq:white}
\end{equation}

\subsection{Novel pose animation under novel illuminations}
\label{subsec:infer}
After optimization, we can get the canonical PBR-aware 2DGS $\mathbf{G}$. To achieve novel pose animations under novel illumination, we first deform the optimized canonical PBR-aware 2DGS to the observation space with the given novel pose which is unseen in the training process following the equation~\ref{eq:lbs}. We then use the environment map of the novel illuminations to build a cubemap. For each viewpoint, we query the cubemap to calculate the PBR color for each Gaussian at observation space with the equation~\ref{eq:pbr}. Finally, we apply the splatting process (equation~\ref{eq:splatting}) to render the avatar at given novel illumination.

%% file: Sec/4_experiments.tex
\section{Experiments}
\label{sec:exp}

\subsection{Evaluation datasets}
To validate the effectiveness of our proposed method,  we use a synthetic dataset (RANA~\cite{iqbal2023rana}) and two real-world datasets (PeopleSnapshot~\cite{peoplesnapshot}, ZJU-MoCap~\cite{peng2021neural,fang2021mirrored}) for performance evaluation.

\textbf{RANA}~\cite{iqbal2023rana}. RANA~\cite{iqbal2023rana} is a synthetic dataset that includes ground-truth albedo, normal, and relighting results at novel pose for evaluation. Each training view is a subject with an A-pose rotating in front of a camera in the scene. The testing set includes the same subject under unknown illumination with random poses. To quantitatively evaluate the physical materials of the reconstructed human, we use 8 subjects from the RANA dataset~\cite{iqbal2023rana} following the same protocols in IntrinsicAvatar~\cite{wang2024intrinsicavatar}. 

\textbf{PeopleSnapshot}~\cite{peoplesnapshot} PeopleSnapshot captures subjects that hold a nearly A-pose and rotate in front of the camera. The subjects are under natural illumination in the real world. We follow previous works~\cite{Lei2024GART, jiang2022instantavatar} and select 4 subjects from the dataset for qualitative evaluation. After optimization, our reconstructed avatar can be rendered in a novel pose under novel illuminations.

\textbf{ZJU-MoCap}~\cite{peng2021neural,fang2021mirrored} In ZJU-MoCap, the subjects captured in the video conduct very complex motions. The shadow effect during the complex motion challenges our physically based rendering. We use the dataset to show that our method can achieve human intrinsic disentangles from a monocular video with complex human motions.

\subsection{Baselines}
We compare our method with two recently proposed methods, IntrinsicAvatar~\cite{wang2024intrinsicavatar} and R4D~\cite{chen2022relighting}.
IntrinsicAvatar~\cite{wang2024intrinsicavatar} is the most recent state-of-the-art method with publicly available training codes for the physically based inversed rendering of clothed humans. We also choose the  Relighting4D (R4D) as our baselines~\cite{chen2022relighting}. Since the original R4D implementation does not utilize the alpha loss, we follow~\cite{wang2024intrinsicavatar} to report a variant of R4D*\footnote{The training code of RelightableAvatar~\cite{xu2024relightable} are not fully released with guidance at the current stage, which we encountered some errors when run the code. They only release one checkpoint on their MobileStage dataset for inference. Thus we do not compare our method with it.} that utilize an alpha loss.

\subsection{Evaluation Metrics}

\begin{figure*}[htp]
    \centering
    \includegraphics[width=\textwidth]{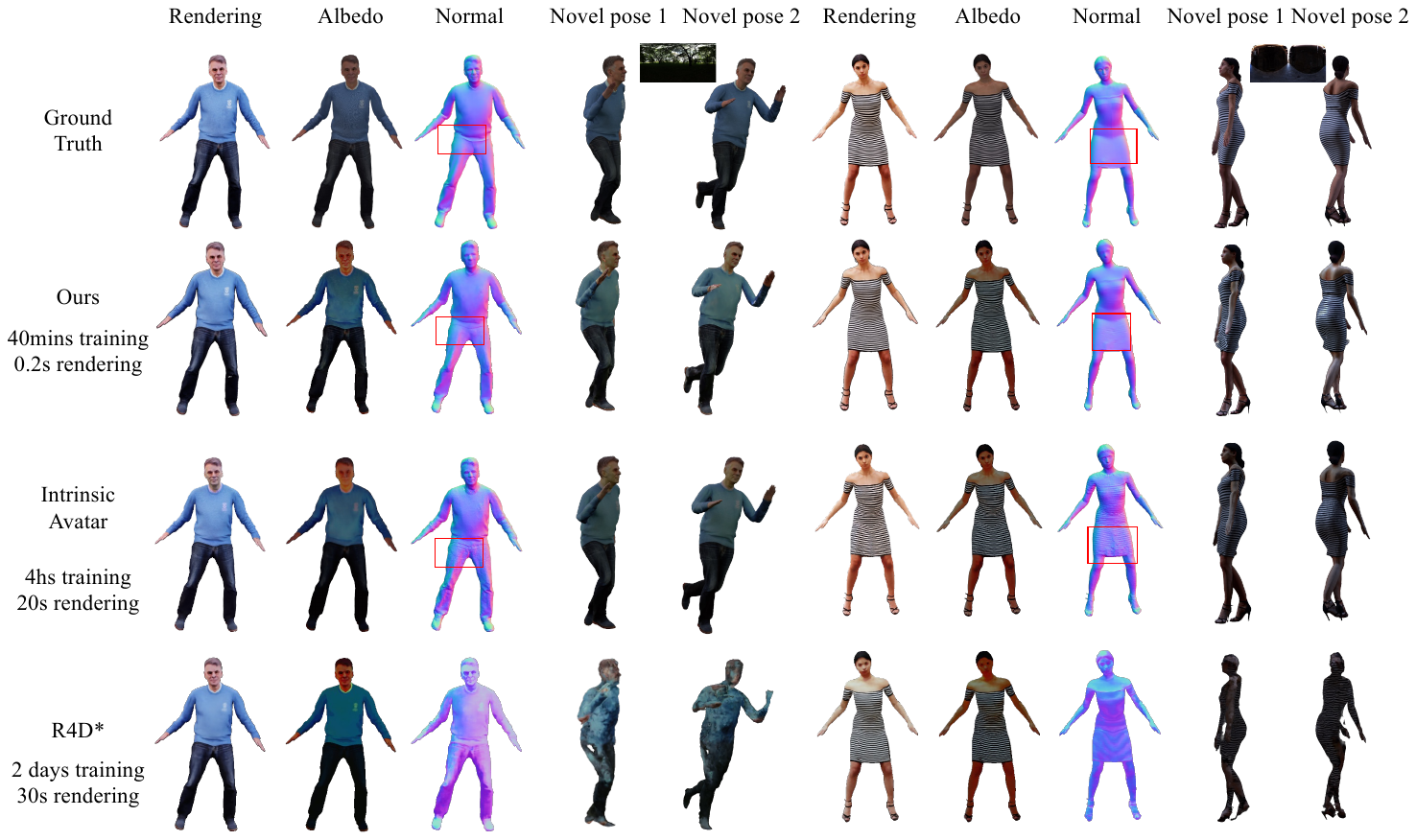}
    \vspace{-8mm}
    \caption{
    Qualitative results on the RANA dataset. We visualize the PBR image, albedo, and normal from the training view. We then relight the clothed human avatars at 2 different novel poses under the novel illuminations.  \mytext{As highlighted by the red rectangles, IntrinsicsAvatar~\cite{wang2024intrinsicavatar} may optimize the fake normal to fit the appearance, while our method provides a more realistic normal for better relighting.}
    }
    \label{fig:rana_exp}
    \vspace{-3mm}
\end{figure*}
\begin{table*}[htp]
\centering
\caption{\textbf{Quantitative comparison to the baselines on the RANA dataset.} The bold font indicates the best performance. The underline indicates the second-best performance. Our method costs minimal training time.}
\resizebox{\linewidth}{!}{
\begin{tabular}{c|c c c |c|c c c|c}
\Xhline{3\arrayrulewidth}
\multirow{2}{*}{Method} &
\multicolumn{3}{c|}{Relighting (novel pose)} &
\multirow{2}{*}{\makecell{Normal \\ Error} $\downarrow$} &
\multicolumn{3}{c|}{Albedo} &
\multirow{2}{*}{Training Time} \\
& 
PSNR $\uparrow$ & SSIM $\uparrow$ & LPIPS $\downarrow$ & &
PSNR $\uparrow$ & SSIM $\uparrow$ & LPIPS $\downarrow$ & \\
\hline
R4D~\cite{chen2022relighting} & 16.41 & 0.8023 & 0.2213  & 42.69$^\circ$ & 18.24 & 0.7780 & 0.2414 & 2 days\\
R4D*~\cite{chen2022relighting} & 18.32 & 0.8132 & 0.1913  & 27.38$^\circ$ & 18.23 & 0.8254 & 0.2043  &  2 days \\
IntrinsicAvatar~\cite{wang2024intrinsicavatar} &  \underline{20.68} & \underline{0.8818} & \underline{0.1139} & \textbf{9.98$^\circ$} & \textbf{22.23} & \underline{0.8862} & \underline{0.1615}  & \underline{4 hours}\\
\hline
Ours  &  \textbf{21.30}  & \textbf{0.8871}  & \textbf{0.1134} & \underline{10.33}$^\circ$ & \underline{21.76}    & \textbf{0.8908} & \textbf{0.1492} &  \textbf{ 40 mins }   
\\
\Xhline{3\arrayrulewidth}
\end{tabular}
}
\label{tab:quantitative}
\vspace{-2mm}
\end{table*}

\begin{figure*}[htp]
    \centering
    \includegraphics[width=\textwidth]{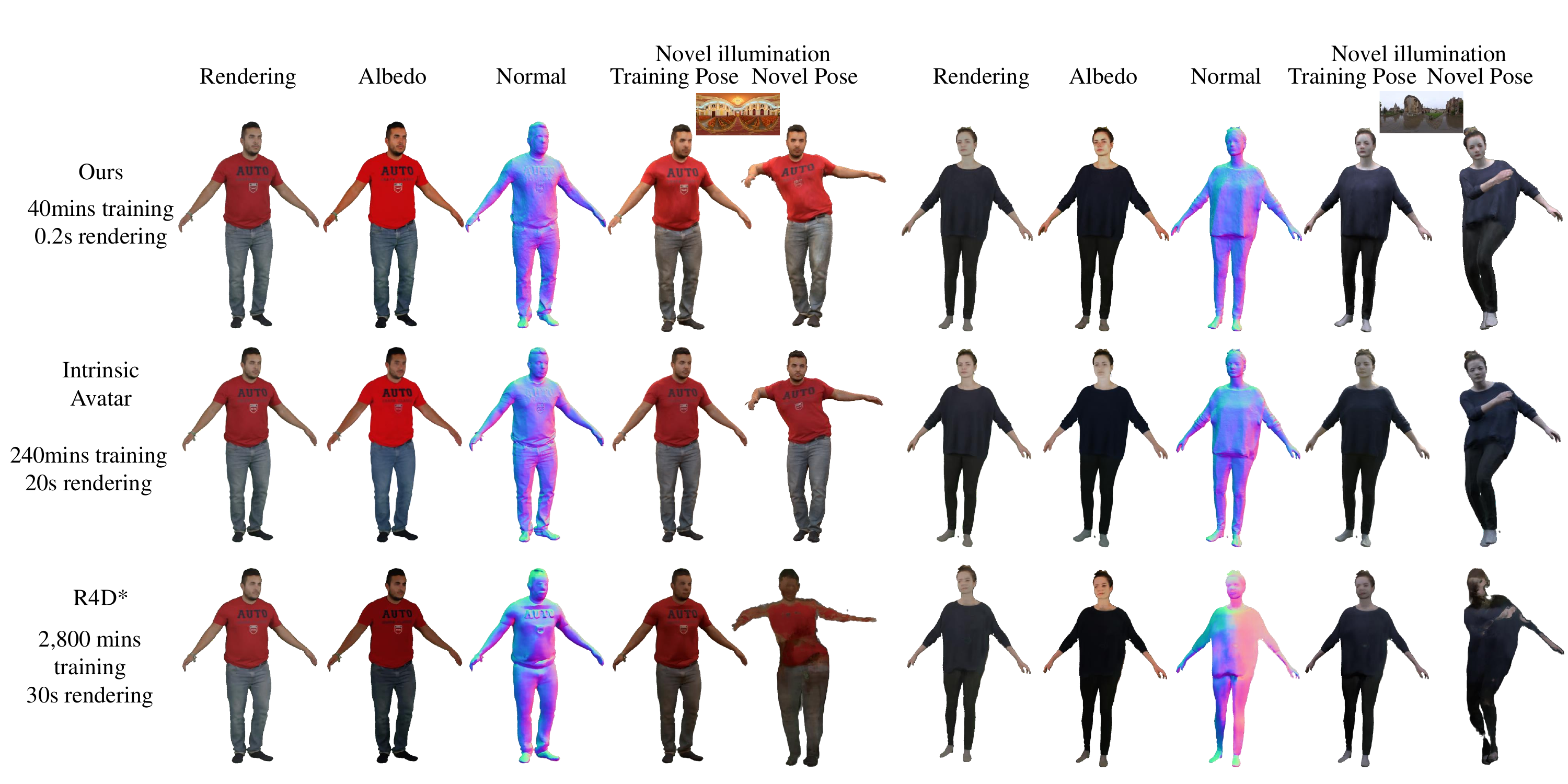}
    \vspace{-8mm}
    \caption{
    Qualitative comparisons on the PeopleSnapshot dataset. We visualize the PBR image, albedo, and normal from the training view. We then relight the clothed human avatars at training pose and novel pose under the novel illumination.  
    }
    \label{fig:people_exp}
    \vspace{-5mm}
\end{figure*}

We use the following metrics for performance evaluation on the synthetic dataset RANA~\cite{iqbal2023rana}. 

\textbf{Albedo: PSNR/SSIM/LPIPS}: We use the image quality metrics to evaluate the albedo estimation from the training view. Since ambiguity exists between the albedo scale and light intensity, we follow previous works~\cite{physg2021, wang2024intrinsicavatar} to align the estimated albedo scale to the ground-truth albedo scale. 

\textbf{Normal Estimation Error}: The metric evaluates the normal errors between the predicted normal and the ground-truth normal from the training view. 

\textbf{Relighting (novel pose): PSNR/SSIM/LPIPS:} We render the clothed human in the novel pose under the novel illumination from the test set. We then evaluate the image quality between the predicted images and the ground-truth.

We only compared our method with baselines qualitatively on the real-world datasets since there is no ground-truth albedo and material information. 
\subsection{Experimental Results}
\begin{figure*}[htp]
    \centering
    \includegraphics[width=\textwidth]{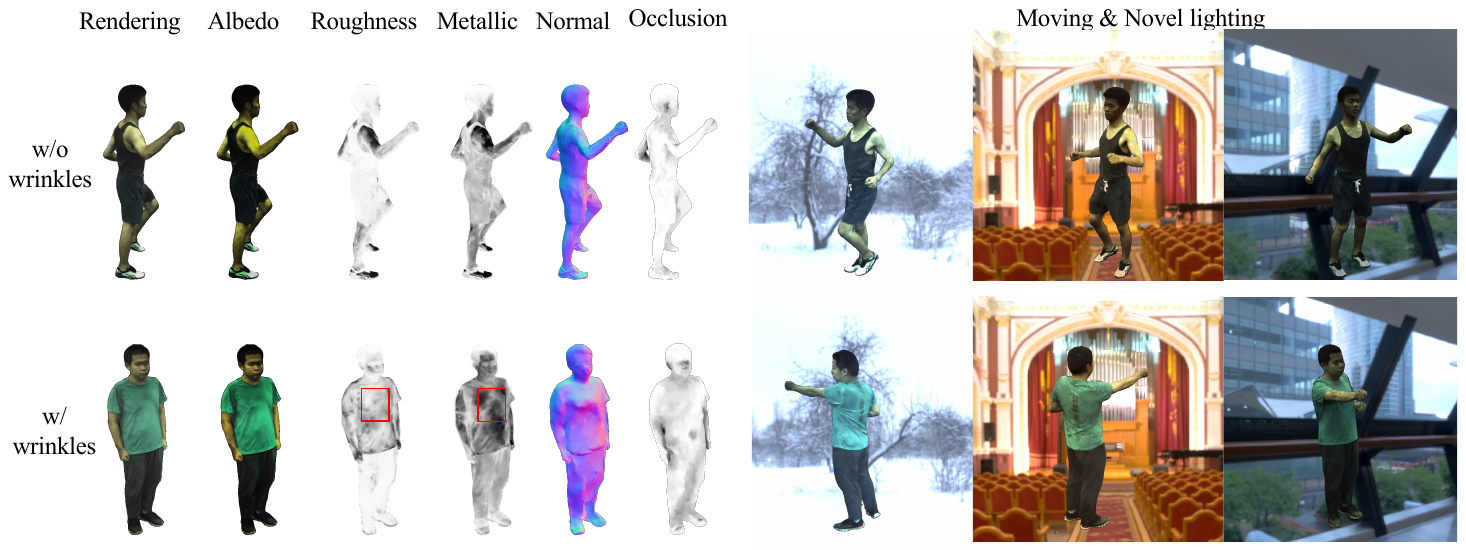}
    \vspace{-5mm}
    \caption{
    Qualitative results on the ZJU-MoCap~\cite{peng2021neural} dataset. In the left 6 columns, we show the rendered image, estimated albedo, roughness, metallic, normal, and the approximated occlusion. In the right 3 columns, we visualize the relighting results under different novel illuminations at different poses. \mytext{Top row: for the avatar with no complex wrinkles on the cloth, our method reconstruct smooth and reasonable physical propertis. Bottom row: for the avatar with complex wrinkles on the cloth, although our method can still work well on most regions, it cannot calculate the physical properties correctly at the winkle area as highlighted by the red rectangles.}
    }
    \label{fig:zju_results}
    \vspace{-5mm}
\end{figure*}
\begin{figure*}[htp]
    \centering
    \includegraphics[width=\textwidth]{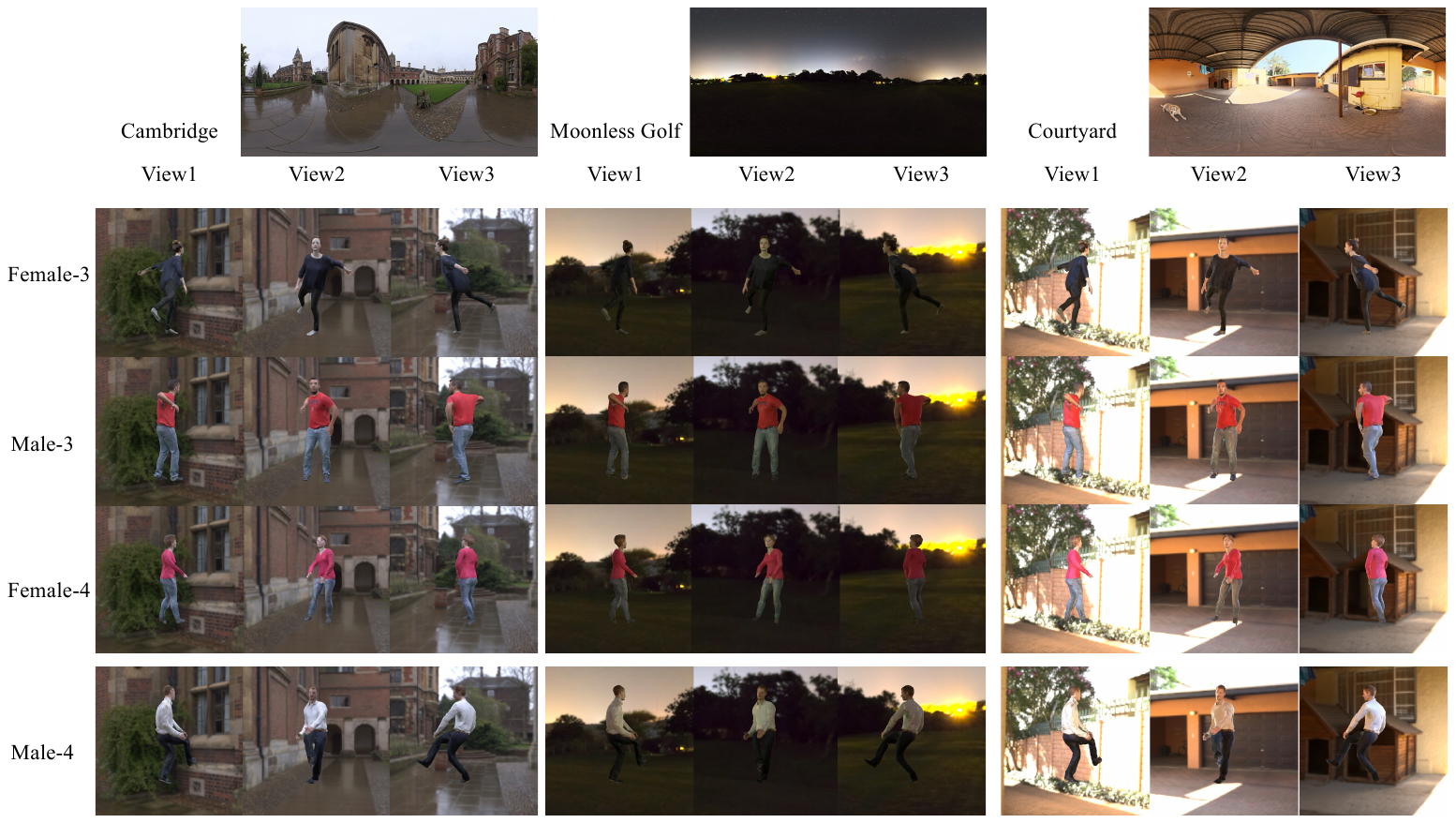}
    \vspace{-10mm}
    \caption{
    Our clothed human avatars in novel views under novel illuminations. The avatars are from the PeopleSnapshot~\cite{peoplesnapshot} dataset. We visualize the avatars from different views with different poses and under different unknown illuminations from~\cite{polyhaven_hdris}.
    }
    \label{fig:more_relight}
    \vspace{-5mm}
\end{figure*}
\noindent\textbf{Training details.} The image reconstruction stage includes 5,000 steps. We apply the normal-depth consistency regularizer after 3,000 steps, from which we can start to align the splat normal to the optimized geometry. We then conduct 3,000 steps to optimize the PBR materials. The overall training step is 8,000 steps, which consumes about 40 minutes on a single Nvidia 3090Ti GPU (with peak 13GB GPU memory usage), which is almost 6 times faster than IntrinsicAvatar.  We set the $\lambda_{\text{NC}}=0.05$ for the normal-consistency loss. The weights in the smooth loss ($\lambda_{\mathbf{a}},\lambda_r, \lambda_m$) are set to be 0.02.  The weight of $\lambda_{\text{white}}$ is set to be 0.1. We follow the learning rate in the original Gaussian attribute in Gaussian Splatting~\cite{kerbl3Dgaussians, Huang2DGS2024}. For the additional parameters of the PBR attribute and the learnable environment light, we set the learning rate the same as that of the spherical harmonics appearance. In the image reconstruction stage, the overall loss is eq.~\ref{eq:1stage}, while in the PBR optimization stage, the overall loss is eq.~\ref{eq:2stage}.

\noindent\textbf{Rendering time.} The rendering time mainly depends on the number of Gaussians. The time cost mainly comes from the occlusion calculation. For PeopleSnapshot~\cite{peoplesnapshot} with $540\times 540$ resolution, there are $\sim50,000$ Gaussians after optimization, the physically based rendering time is 0.2 seconds per image, which is faster than all previous neural-rendering based works (20s per image as shown in the Fig.~\ref{fig:rana_exp}).

\subsubsection{Comparison on synthetic dataset}


We compare with the baselines in Tab.\ref{tab:quantitative}~\footnote{ We re-run the baselines~\cite{chen2022relighting, wang2024intrinsicavatar} with the official implementation provided by the authors: \\ \href{https://github.com/FrozenBurning/Relighting4D}{https://github.com/FrozenBurning/Relighting4D}, 
\\
\href{https://github.com/taconite/IntrinsicAvatar}{https://github.com/taconite/IntrinsicAvatar} on our dataset splits and report the reproduced both the quantitative and qualitative results. Our reported relighting results are higher than those reported in previous works as we employ a similar processing method to albedo by dilating the foreground outward by 100 pixels.} and Fig.\ref{fig:rana_exp}. We can observe that R4D*~\cite{chen2022relighting} cannot reconstruct the surface geometry accurately. Since such a method relies on the time-conditioned embedding during optimization, it cannot generalize to the novel poses. \mytext{Moreover, its implicit representation lacks detail.} While Intrinsic Avatar~\cite{wang2024intrinsicavatar} can generalize to the novel pose and its multi-level Hash-Grids~\cite{mueller2022instant} representation provide much more details, the albedo details (in the man's face on the left of Fig.~\ref{fig:people_exp}) are still missing. For the female on the right of Fig.~\ref{fig:people_exp}, the dress surface reconstructed by ~\cite{wang2024intrinsicavatar} tends to overfit the texture, while our methods can reconstruct the smooth surface for the dress. \mytext{Furthermore, our method can bring high light during relighting, which brings superiority for relighting (novel pose) as shown in the relighting results on the back of the female in the Fig.~\ref{fig:rana_exp}}. The quantitative results in  Tab.~\ref{tab:quantitative} also demonstrate this. Compared with R4D and R4D*, our methods reduce $75.8\%$ and $62.6\%$ normal errors and significantly improve the PSNR of estimated albedo by $19.2\%$. Compared with IntrinsicAvatar, our methods achieve comparable estimated albedo quality (a lower PSNR (0.46 dB), but a higher SSIM and LPIPS) and more accurate relighting quality (a higher PSNR (0.62 dB)). Further, compared with SOTA methods, our methods can achieve a $5\times$ faster in training speed and a $100\times$ faster in rendering speed.

\subsubsection{Comparison on real-world dataset}

We also qualitatively compare our method with the baselines on the dataset of nearly A-pose in the real world in Fig.~\ref{fig:people_exp}. The novel poses are selected from the AIST dataset~\cite{li2021learn}. We can observe that the albedo estimation of R4D*~\cite{chen2022relighting} fails to recover the plausible solution, its geometry is inaccurate. The relighting results at both the training and test poses are unsuitable for environmental light. Compared with Intrinsic Avatar~\cite{wang2024intrinsicavatar}, our methods can produce more plausible relighting results in addition to $5\times$ faster in training speed and $100\times$ faster in rendering speed.
\subsubsection{Comparision with non-PBR 3DGS-based baseline}

\begin{figure}[htp]
    \centering
    \vspace{-4mm}
    \includegraphics[width=0.5\textwidth]{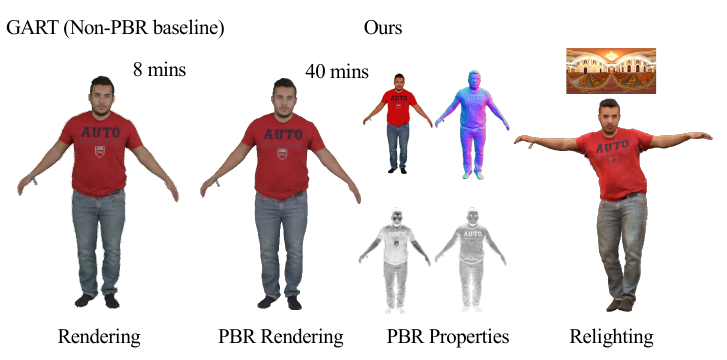}
     \vspace{-8mm}
    \caption{
    \mytext{\textbf{Comparisions with Non-PBR baseline GART~\cite{Lei2024GART} }. Our method achieves PBR physical properties including albedo, normal, metallics and roughness, while GART can only support radiance fields reconstruction.}}
  \vspace{-5mm}
    \label{fig:with_gart}
\end{figure}

\mytext{
We compared our method with GART ~\cite{Lei2024GART}, a non-PBR baseline for human reconstruction in the Fig.~\ref{fig:with_gart}. Although our method consumes more training time, it can successfully reconstruct all the PBR properties of the clothed avatar and support the relighting under novel illuminations. Moreover, as demonstrated in the Tab.~\ref{tab:ablate_clothed}, the occlusion calculation is the main bottleneck for time consumption compared with the non-PBR baseline. Our training speed improvements benefit from our proposed occlusion calculation.
}

\subsubsection{PBR properties estimation results on the real-world datasets with complex motions}

We follow the setting in the InstantNVR~\cite{instant_nvr} and choose view 4 from ZJU-MoCap as the training view. We visualize the intrinsic estimation results in Fig.~\ref{fig:zju_results}. Our methods can produce plausible BRDF estimation results. We also render the clothed human at novel illumination and get well-reproduced results. We also visualize the approximated occlusion calculated from the template mesh to show the effectiveness of our proposed human avatar occlusion approximation strategy. We can observe from Fig.~\ref{fig:zju_results} that our method can achieve a plausible PBR properties estimation from a monocular video with complex human motions, and can also re-render them with an appearance suitable to the environment lighting. \mytext{Moreover, since our proposed occlusion calculation strategy is based on the SMPL human model, it cannot calculate the occlusion correctly at the complex geometric region, therefore leading to the wrong physical properties at the winkle area as highlighted by the red rectangles in Fig.~\ref{fig:zju_results}. We also provide a comprehensive discussion about this at the Section ~\ref{sec:conc}.}

\subsubsection{Reposed clothed human avatars in novel views under novel illuminations}

We select the motion sequence from the CAPE dataset~\cite{CAPE_CVPR, ClothCap} and repose our reconstructed clothed human avatars. We then select 3 different views and re-render each avatar under 3 different environment lighting conditions. The results in Fig.~\ref{fig:more_relight} demonstrate the robustness of our methods. The reconstructed avatars from our methods can generalize well to the out-of-distribution pose under totally novel illuminations while maintaining 3D consistency. We provide a video in our supplementary to provide a temporally consistent visualization.

\begin{figure*}[htp]
    \centering
    \includegraphics[width=\textwidth]{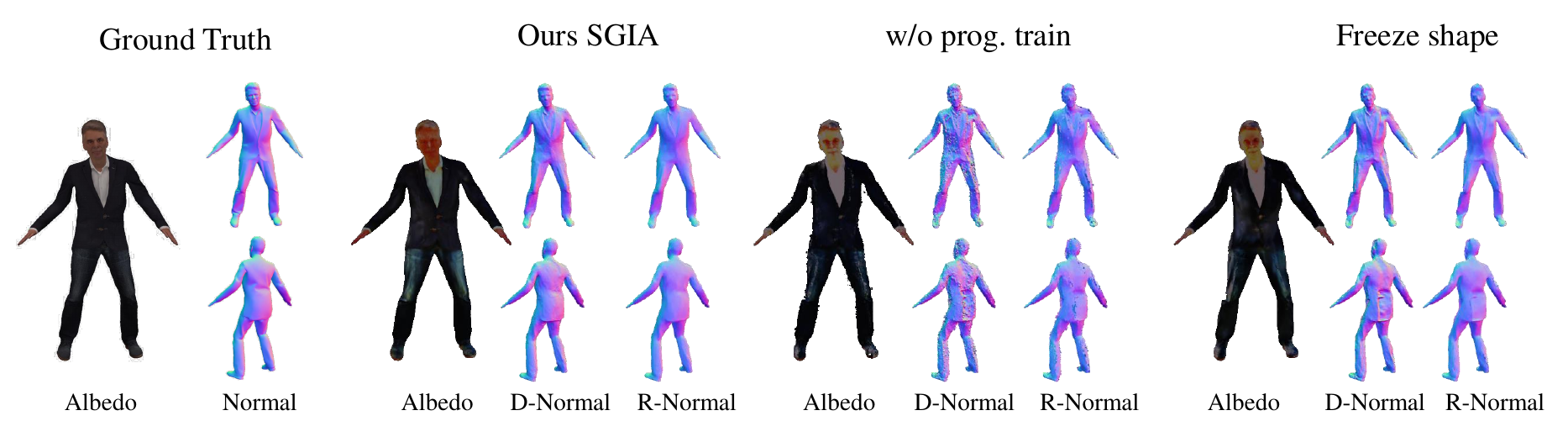}
    \caption{
    \textbf{Ablation studies of different training strategies.} We visualize the Albedo estimation, D-Normal (normal from depth points), and the R-Normal (rendered normal of each splat). 
  }
  \vspace{-5mm}
    \label{fig:ablation_training}
\end{figure*}


\begin{table}[tp]
\centering
\caption{\textbf{Ablation study of our proposed training strategy}}
\resizebox{\linewidth}{!}{
\begin{tabular}{c|c c c |c}
\Xhline{3\arrayrulewidth}
\multirow{2}{*}{Method} &
\multicolumn{3}{c|}{Albedo} &
\multirow{2}{*}{\makecell{Normal \\ Error} $\downarrow$} \\
& PSNR $\uparrow$ & SSIM $\uparrow$ & LPIPS $\downarrow$ & \\
\hline
Freeze shape & 18.78 & 0.7661 & 0.2581 & 21.51$^\circ$ \\ 
w/o prog. train & 20.21 & 0.8247 & 0.1977 & 12.89$^\circ$\\
Ours SGIA & 23.17 & 0.8839 & 0.1701 & 11.31$^\circ$\\ 
\Xhline{3\arrayrulewidth}
\end{tabular}
}
\label{tab:ablation_training}
\end{table}
\subsection{Ablation studies}

We conduct ablation studies on different representations (3DGS or 2DGS), different normal consistency regularization($\ell_1$-loss or cosine similarity loss), the effectiveness of occlusion calculation, the effectiveness of the occlusion approximation strategies, the PBR regularization loss, and our progressive training strategy.

\subsubsection{Training strategy}

\begin{table}[tb]
\centering
\caption{\textbf{Ablation study of different primitive representation}}
\resizebox{\linewidth}{!}{
\begin{tabular}{c|c c c |c}
\Xhline{3\arrayrulewidth}
\multirow{2}{*}{Method} &
\multicolumn{3}{c|}{Albedo} &
\multirow{2}{*}{\makecell{Normal \\ Error} $\downarrow$} \\
& PSNR $\uparrow$ & SSIM $\uparrow$ & LPIPS $\downarrow$ & \\
\hline
PBR-aware 3DGS  & 22.07 & 0.8739 & 0.1586 & 12.33$^\circ$\\
PBR-aware 2DGS & 22.43 & 0.8923 & 0.1545 & 11.71$^\circ$\\ 
\Xhline{3\arrayrulewidth}
\end{tabular}
}
\label{tab:ablation_primitive}
\end{table}
\begin{figure}[tb]
    \centering
    \includegraphics[width=0.5\textwidth]{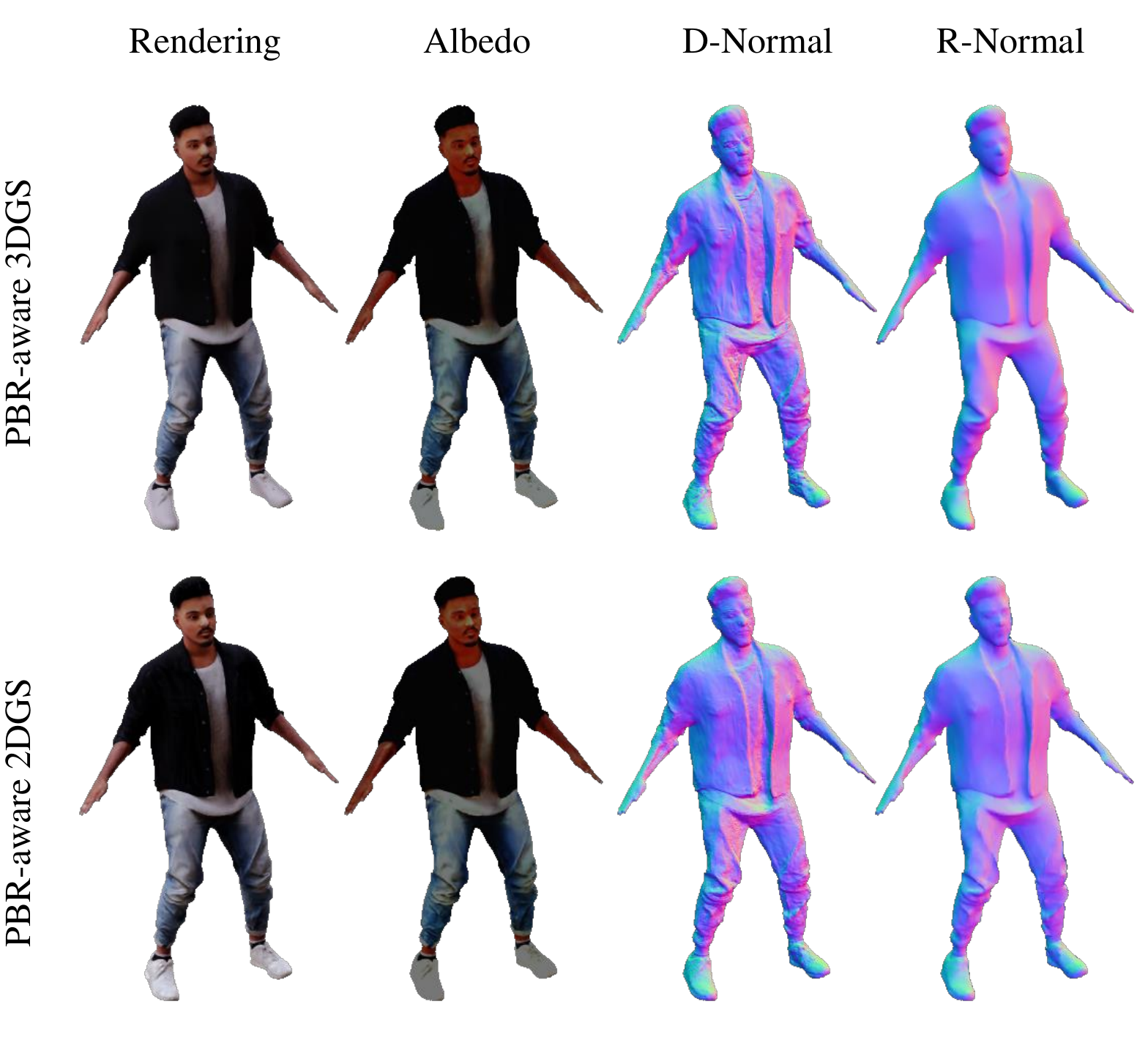}
    \vspace{-10mm}
    \caption{
    Ablation studies of different primitive choices. We visualize the rendering image, albedo, D-Normal (normal from depth points), and the R-Normal (rendered normal of each splat).
     }
     \vspace{-3mm}
    \label{fig:ablation_primitive}
\end{figure}

\noindent We first compare different training strategies for the PBR optimization stage in the Tab.~\ref{tab:ablation_training} and the Fig.~\ref{fig:ablation_training}. "Freeze shape" means that we choose the optimization strategy of GS-IR~\cite{liang2023gs}, which freeze the shape after image reconstruction stage and only optimize the PBR materials of each Gaussian. "w/o progressive" means that we do not utilize the strategy in (equation~\ref{eq:progressive}) and directly utilize the normal-consistency loss (equation~\ref{eq:lnc}) from the first stage. We can observe from the Fig.~\ref{fig:ablation_training} that if we utilize the "freeze shape", the geometry of Gaussian is fixed. \getext{The initial geometry from the first stage does not improve anything. Without our progressive training, the Gaussian normals would always align with the inaccurate geometry from the first stage, hindering geometry optimization in the PBR stage.} With our proposed progressive training strategy, both the normal of each splat and the geometry of the avatar can be aligned consistently to the accurate geometry. Moreover, the inaccurate geometry also leads to inaccurate albedo estimation. The quantitative results in Tab.~\ref{tab:ablation_training} further demonstrate these.

\subsubsection{Primitive representation}


We ablate different choices of primitive representation in Tab.~\ref{tab:ablation_primitive} and Fig.~\ref{fig:ablation_primitive}. Since original 3DGS~\cite{kerbl3Dgaussians} do not have an attribute for splat orientation, we follow the previous relightable 3DGS~\cite{liang2023gs, R3DG2023} to assign a learnable normal vector for each Gaussian and keep all the other \getext{strategies} unchanged. We can observe from Fig.~\ref{fig:ablation_primitive} that with 3DGS as a primitive representation, though the splat normal (R-normal) can be very smooth, there exists a lot of uneven region in the Depth (D-normal). The R-normal and the D-Normal are not consistent. However, with the 2DGS representation, the R-Normal and the D-Normal can be both consistent and accurate. The Normal Error in Tab.~\ref{tab:ablation_primitive} also demonstrates this quantitatively.

\subsubsection{Normal-consistency loss}

\begin{table*}[htp]
\centering
\caption{\textbf{Ablation study of different consistency loss}}
\resizebox{0.8\linewidth}{!}{
\begin{tabular}{c|c c c |c|c c c}
\Xhline{3\arrayrulewidth}
\multirow{2}{*}{Method} &
\multicolumn{3}{c|}{Albedo} &
\multirow{2}{*}{\makecell{Normal \\ Error} $\downarrow$} &
\multicolumn{3}{c}{Relighting (novel pose)} \\
& 
PSNR $\uparrow$ & SSIM $\uparrow$ & LPIPS $\downarrow$ & &
PSNR $\uparrow$ & SSIM $\uparrow$ & LPIPS $\downarrow$ \\
\hline
$\ell_1$-loss & 21.76 & 0.8798 & 0.1746 & 20.33$^\circ$  & 15.87 & 0.7844 & 0.2162 \\
cosine-loss & 22.43 & 0.8923 & 0.1545 & 11.71$^\circ$ & 19.77 & 0.8642 & 0.1317 \\  
\Xhline{3\arrayrulewidth}
\end{tabular}
}
\label{tab:ablation_consistency}
\vspace{-4mm}
\end{table*}
\begin{figure}[htp]
    \centering
    \includegraphics[width=0.46\textwidth]{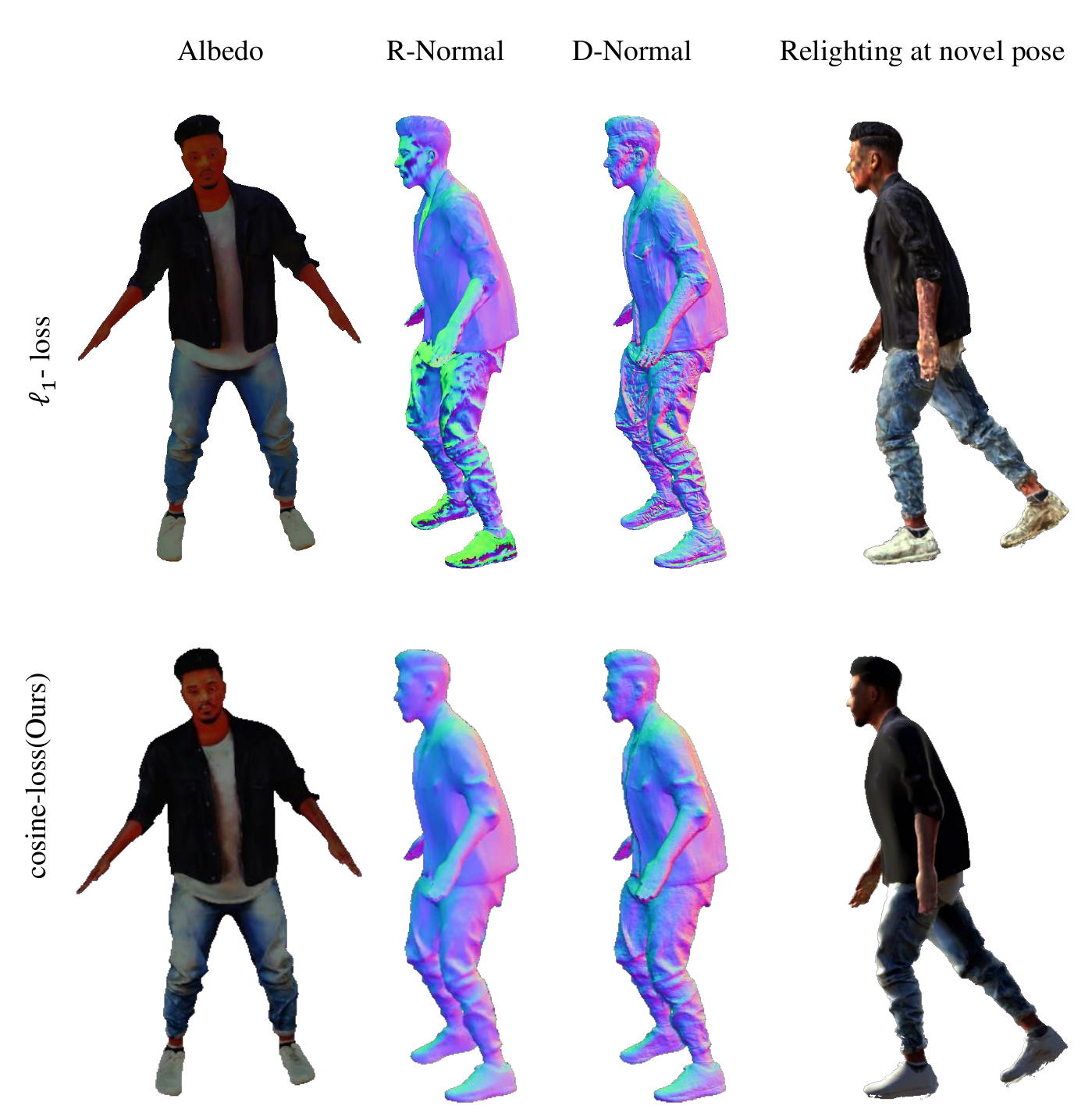}
    \vspace{-5mm}
    \caption{
    Ablation studies of different normal-consistency loss choices. We visualize the albedo, the D-Normal (normal from depth points), the R-Normal (rendered normal of each splat), and the relighting results with novel poses under a novel illumination.
     }
    \label{fig:ablation_consistency}
\end{figure}

\noindent We ablate the normal consistency loss in Tab~\ref{tab:ablation_consistency} and Fig.~\ref{fig:ablation_consistency}. "$\ell_1$-loss" means that we follow the previous relightable 3DGS~\cite{R3DG2023, liang2023gs} and utilize the $\ell_1$-loss to minimize the difference of the linear blended normal and the normal from depth points. \getext{When we keep the other factors unchanged and use the $\ell_1$-loss, the splat normals cannot align well with the D-Normals. This inaccurate splat normal also leads to poor relighting results, as shown in Fig.}~\ref{fig:ablation_consistency}. The quantitative results \getext{in the Tab~\ref{tab:ablation_consistency} further support this observation.}

\begin{figure}[htp]
    \centering
    \includegraphics[width=0.49\textwidth]{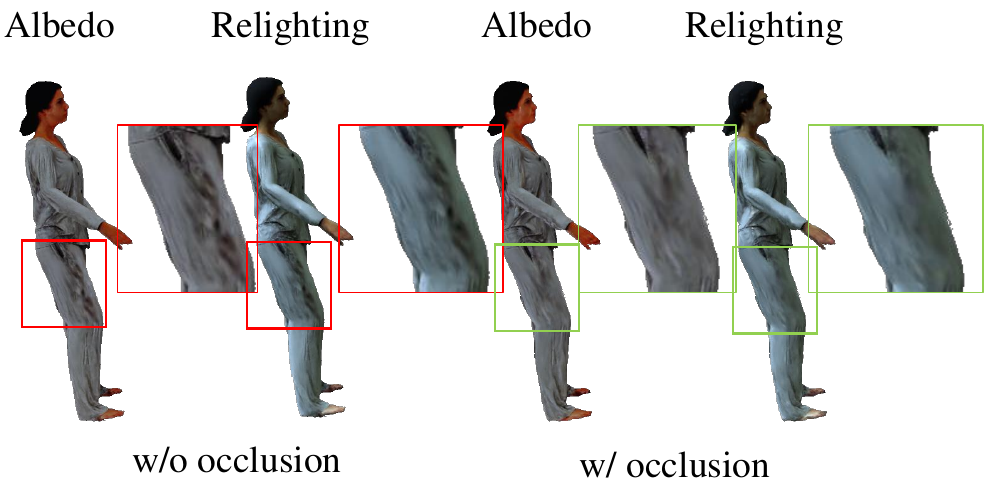}
    \vspace{-5mm}
    \caption{
    Ablation study of our proposed occlusion calculation. We visualize the Albedo and Relighting results under novel illuminations, both at novel poses. The red rectangle indicates the coupled occlusion information in albedo. The green rectangle indicates the decoupled occlusion information from the albedo. 
     }
    \vspace{-2mm}
    \label{fig:ablate_occlusion}
\end{figure}
\begin{table}[htp]
\centering
\caption{\textbf{Ablation study of the effectiveness of the occlusion calculation}}
\resizebox{\linewidth}{!}{
\begin{tabular}{c|c c c |c}
\Xhline{3\arrayrulewidth}
\multirow{2}{*}{Method} &
\multicolumn{3}{c|}{Albedo} &
\multirow{2}{*}{\makecell{Normal \\ Error} $\downarrow$} \\
& PSNR $\uparrow$ & SSIM $\uparrow$ & LPIPS $\downarrow$ & \\
\hline
w/o occlusion  & 21.08 & 0.8616 & 0.1711 & 11.87$^\circ$\\
w/ occlusion & 21.31 & 0.8701 & 0.1702 & 10.69$^\circ$\\
\Xhline{3\arrayrulewidth}
\end{tabular}
}
\vspace{-5mm}
\label{tab:occlusion}
\end{table}

\subsubsection{Occlusion calculation}

We ablate our proposed occlusion calculation in Tab.~\ref{tab:occlusion} and Fig.~\ref{fig:ablate_occlusion}. We can observe that the shadow information may be mixed in the albedo without the occlusion calculation during the optimization stage. With the occlusion calculation, our methods successfully disentangle the occlusion information from the albedo. The quantitative results in Tab.~\ref{tab:occlusion} also prove the effectiveness of decoupling the shadow effect and the PBR properties of our occlusion approximation strategy.


\subsubsection{Occlusion calculation from template mesh or clothed mesh}
\label{subsub:occ}
\begin{figure}[htp]
    \centering
    \includegraphics[width=0.5\textwidth]{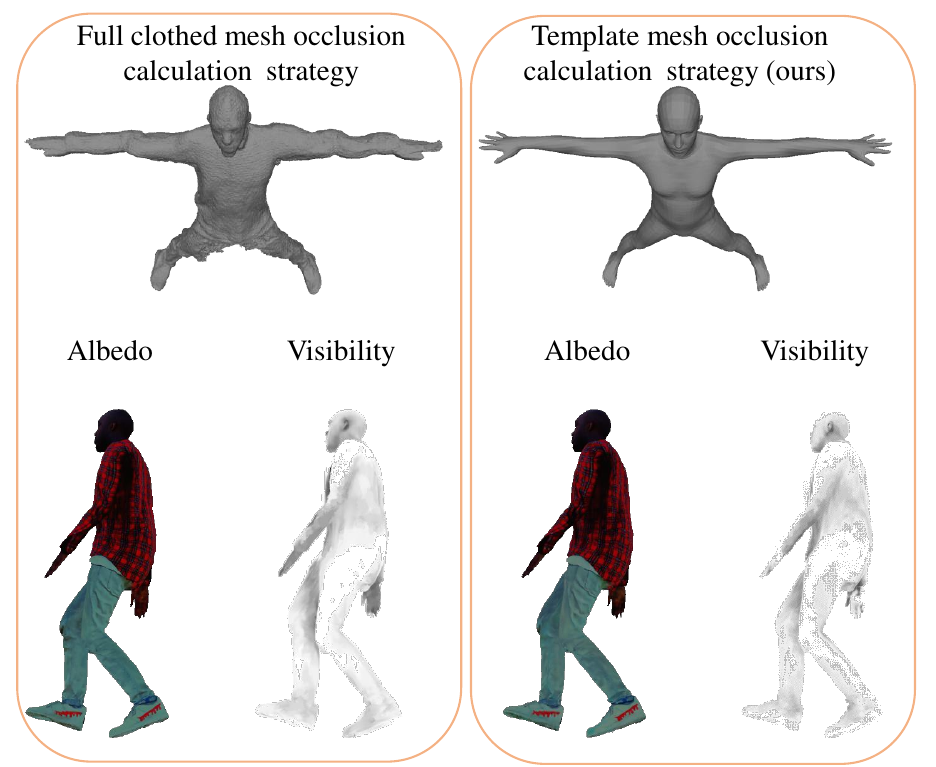}
    \vspace{-5mm}
    \caption{
    Ablation studies of different occlusion calculation strategies. At the top of the figure, we visualize different canonical meshes used in different occlusion calculation strategies. At the bottom of the figure, we visualize the albedo and the visibility map of different occlusion calculation strategies both at novel poses.   
     }
     \vspace{-2mm}
    \label{fig:ablate_clothed}
\end{figure}

We ablate our approximation of the occlusion calculation from the template mesh in Tab.~\ref{tab:ablate_clothed} and Fig.~\ref{fig:ablate_clothed}. For the fully clothed mesh case, we follow 2DGS~\cite{Huang2DGS2024} and use the TSDF Fusion~\cite{zhou2018open3d} to extract the mesh of the clothed human avatar at the canonical pose. We query the points by using tri-linear interpolation from the canonical skinning weight voxel to get the skinning weight for each canonical vertex. We keep other regularization terms the same as those in 2DGS and optimize the same steps. Tab.~\ref{tab:ablate_clothed} \getext{ shows the estimated PBR properties with occlusion from the template mesh have a minor gap to the fully clothed mesh. But rendering time with the clothed mesh is} 10$\times$ longer \getext{, as it has 20x more vertices and faces.} Fig.~\ref{fig:ablate_clothed} \getext{shows the visibility map and estimated albedo are similar, demonstrating our occlusion approximation achieves faster rendering without sacrificing PBR accuracy.}

\begin{table}[htp]
\centering
\caption{\textbf{Ablation study of the occlusion calculation from the template mesh or the clothed mesh}}
\resizebox{\linewidth}{!}{
\begin{tabular}{c|c c c |c}
\Xhline{3\arrayrulewidth}
\multirow{2}{*}{Method} &
\multicolumn{3}{c|}{Albedo} &
\multirow{2}{*}{\makecell{Rendering \\ Time} $\downarrow$} \\
& PSNR $\uparrow$ & SSIM $\uparrow$ & LPIPS $\downarrow$ & \\
\hline
Full clothed mesh  & 22.01 & 0.8831 & 0.1267 & 2s\\
Template mesh (Ours)& 21.93 & 0.8827 & 0.1259 & 0.2s\\
\Xhline{3\arrayrulewidth}
\end{tabular}
}
\vspace{-4mm}
\label{tab:ablate_clothed}
\end{table}
\subsubsection{PBR regularization}
\begin{figure}[htp]
    \centering
    \includegraphics[width=0.5\textwidth]{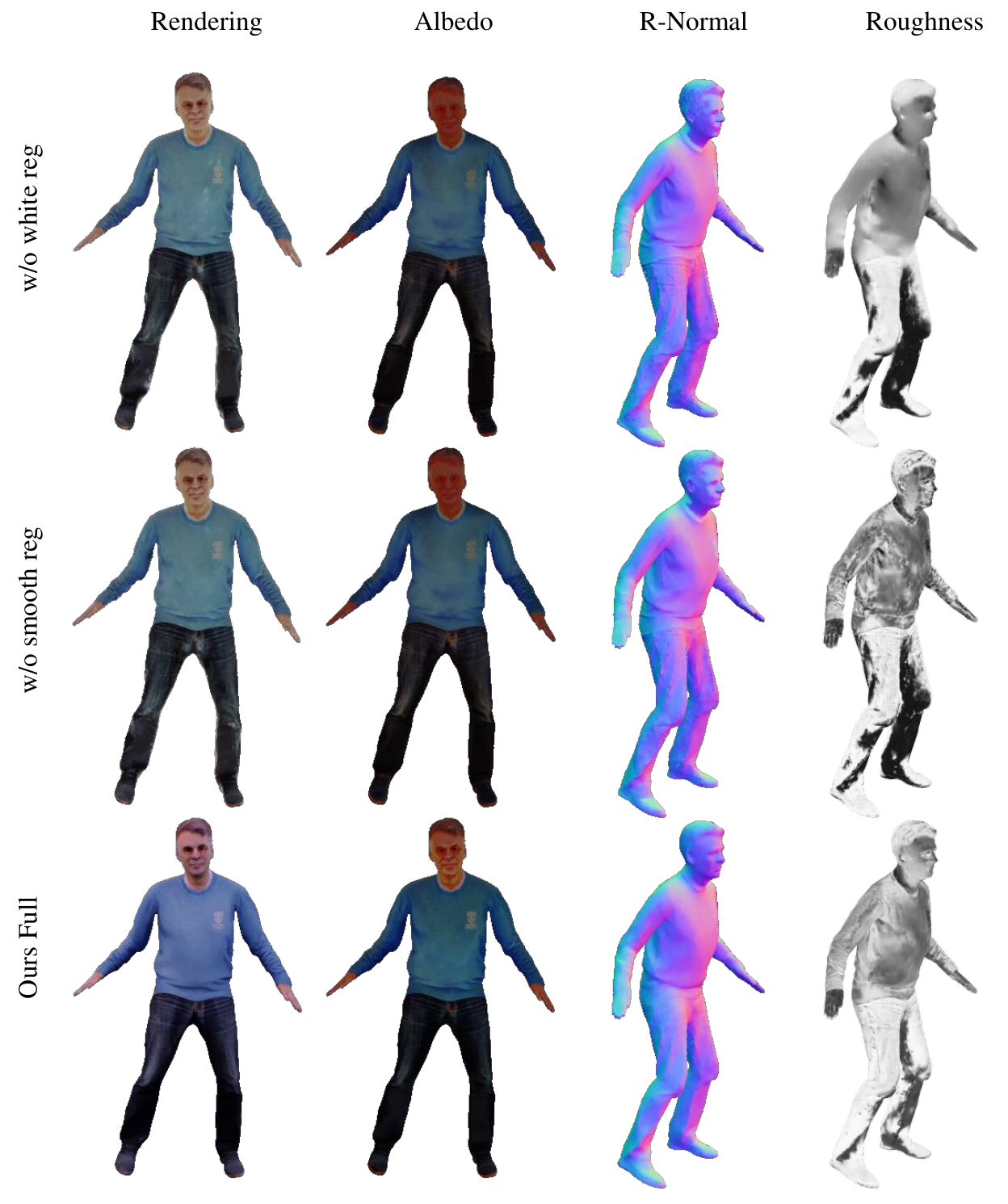}
    \caption{
    Ablation study of PBR Regularization. We visualize the rendering image, albedo, R-Normal (rendered normal of each splat), and the roughness.
     }
     \vspace{-5mm}
    \label{fig:ablate_regularization}
\end{figure}

We ablate our PBR regularization choices in the Tab.~\ref{tab:regularization} and Fig.~\ref{fig:ablate_regularization}. "w/o. white reg" means that we do not utilize the white light regularization in Equation~\ref{eq:white}. "w/o. smooth reg"  means that we do not utilize the smooth regularization in Equation~\ref{eq:smooth}. \getext{ From Fig.~\ref{fig:ablate_regularization}, we can see that without light regularization, the albedo and splat normals are noisy.  Without smoothness regularization, the albedo lacks details, and the splat normals and roughness are chaotic, with inconsistent roughness even in regions of the same material. This leads to PBR renders that don't closely match the ground truth. The quantitative results in Tab~\ref{tab:regularization} also show the effectiveness of these regularization terms.}


\begin{table}[htp]
\centering
\caption{\textbf{Ablation study of PBR regularization}}
\resizebox{\linewidth}{!}{
\begin{tabular}{c|c c c |c}
\Xhline{3\arrayrulewidth}
\multirow{2}{*}{Method} &
\multicolumn{3}{c|}{Albedo} &
\multirow{2}{*}{\makecell{Rendering \\ Time} $\downarrow$} \\
& PSNR $\uparrow$ & SSIM $\uparrow$ & LPIPS $\downarrow$ & \\
\hline
w/o white reg & 22.05 & 0.9045 & 0.1761 & 10.71$^\circ$\\
w/o smooth reg  & 21.83 & 0.9046 & 0.1863 & 10.93$^\circ$\\
Ours Full & 22.44 & 0.8874 & 0.1758 & 9.4$^\circ$\\ 
\Xhline{3\arrayrulewidth}
\end{tabular}
}
\vspace{-5mm}
\label{tab:regularization}
\end{table}



%% file: Sec/6_discussion.tex
\section{Limitation And Future Work} 

While our method efficiently reconstructs the PBR properties of clothed human avatars from monocular videos, it is important to acknowledge its limitations. 

1). One significant limitation is the lack of detailed facial expression capture. Our approach does not leverage expression capture techniques, resulting in less detailed facial animations for the avatars. Future work could enhance realism and expressiveness by integrating facial expression techniques with our full-body gesture reconstruction, as explored in studies such as~\cite{Xu2024, qian2023gaussianavatars, saito2024relightable}. A promising direction is to combine the expressive Gaussian Avatar representation with our full PBR representation pipeline to achieve both relightable and expressive human avatar reconstruction. 

\mytext{2). Another severe limitation comes from the SMPL model. Although the SMPL model effectively achieves a highly efficient occlusion approximation for successfully estimating physical properties, it cannot model wrinkles, loose clothes, or other detailed geometric deformations caused by clothes since it only models the unclothed human mesh. Exploring other detailed human models~\cite{cai2023smplerx, yin2025smplest} for the occlusion approximation is important to achieve a detailed occlusion calculation.}

\mytext{3).  Lastly, since our method focuses on the reconstruction of the PBR properties from monocular videos, our current PBR-based inverse rendering can only provide one possible solution as most previous works~\cite{wang2024intrinsicavatar, chen2022relighting}. Scale ambiguity may exist between the reconstructed PBR properties and the unknown illumination light intensity. Incorporating the diffusion-based intrinsic estimation method~\cite{careagaColorful, careagaIntrinsic, chen2024intrinsicanything} is a promising research direction to solve the scale ambiguity issue. }

%% file: Sec/5_conclusion.tex
\section{Conclusion} 
\label{sec:conc}
In this paper, we propose SGIA, a relightable clothed human avatar representation that can achieve fast reconstruction of the PBR properties from monocular videos. In contrast to all previous NeRF-based works that rely on Monte-Carlo sampling to integrate the in-going radiance, our methods can be easily equipped with pre-integration to achieve efficient lighting computation. We approximate the occlusion by ray-casting from template mesh to balance accuracy and speed. Furthermore, to achieve accurate geometry reconstruction guided by PBR, we propose a simple but effective progressive optimization strategy that improves the geometry and then aligns the splat normal with the geometry. Overall, our methods can achieve five times the training speed and 100 times the rendering speed compared with previous SOTA approaches while maintaining high-quality PBR properties estimation and realistic re-lighting results.


%% file: main.bbl
\begin{thebibliography}{100}
\providecommand{\url}[1]{#1}
\csname url@samestyle\endcsname
\providecommand{\newblock}{\relax}
\providecommand{\bibinfo}[2]{#2}
\providecommand{\BIBentrySTDinterwordspacing}{\spaceskip=0pt\relax}
\providecommand{\BIBentryALTinterwordstretchfactor}{4}
\providecommand{\BIBentryALTinterwordspacing}{\spaceskip=\fontdimen2\font plus
\BIBentryALTinterwordstretchfactor\fontdimen3\font minus \fontdimen4\font\relax}
\providecommand{\BIBforeignlanguage}[2]{{%
\expandafter\ifx\csname l@#1\endcsname\relax
\typeout{** WARNING: IEEEtran.bst: No hyphenation pattern has been}%
\typeout{** loaded for the language `#1'. Using the pattern for}%
\typeout{** the default language instead.}%
\else
\language=\csname l@#1\endcsname
\fi
#2}}
\providecommand{\BIBdecl}{\relax}
\BIBdecl

\bibitem{Barron2014}
J.~T. Barron and J.~Malik, ``Shape, illumination, and reflectance from shading,'' \emph{IEEE Transactions on Pattern Analysis and Machine Intelligence}, vol.~37, no.~8, pp. 1670--1687, 2014.

\bibitem{Li2018}
Z.~Li, Z.~Xu, R.~Ramamoorthi, K.~Sunkavalli, and M.~Chandraker, ``Learning to reconstruct shape and spatially-varying reflectance from a single image,'' \emph{ACM Transactions on Graphics}, vol.~37, no.~6, pp. 269:1--269:11, 2018.

\bibitem{Li2020}
Z.~Li, M.~Shafiei, R.~Ramamoorthi, K.~Sunkavalli, and M.~Chandraker, ``Inverse rendering for complex indoor scenes: Shape, spatially-varying lighting and svbrdf from a single image,'' in \emph{Proceedings of the IEEE Conference on Computer Vision and Pattern Recognition (CVPR)}, 2020.

\bibitem{Lichy2021}
D.~Lichy, J.~Wu, S.~Sengupta, and D.~W. Jacobs, ``Shape and material capture at home,'' in \emph{Proceedings of the IEEE Conference on Computer Vision and Pattern Recognition (CVPR)}, 2021.

\bibitem{Sang2020}
S.~Sang and M.~Chandraker, ``Single-shot neural relighting and svbrdf estimation,'' in \emph{Proceedings of the European Conference on Computer Vision (ECCV)}, 2020.

\bibitem{Wei2020}
X.~Wei, G.~Chen, Y.~Dong, S.~Lin, and X.~Tong, ``Object-based illumination estimation with rendering-aware neural networks,'' in \emph{Proceedings of the European Conference on Computer Vision (ECCV)}, 2020.

\bibitem{Yu2019}
Y.~Yu and W.~A.~P. Smith, ``Inverserendernet: Learning single image inverse rendering,'' in \emph{Proceedings of the IEEE Conference on Computer Vision and Pattern Recognition (CVPR)}, 2019.

\bibitem{Goel2020}
P.~Goel, L.~Cohen, J.~Guesman, V.~Thamizharasan, J.~Tompkin, and D.~Ritchie, ``Shape from tracing: Towards reconstructing 3d object geometry and svbrdf material from images via differentiable path tracing,'' in \emph{Proceedings of the International Conference on 3D Vision (3DV)}, 2020.

\bibitem{Guo2019}
K.~Guo, P.~Lincoln, P.~L. Davidson, J.~Busch, X.~Yu, M.~Whalen, G.~Harvey, S.~Orts-Escolano, R.~Pandey, J.~Dourgarian, D.~Tang, A.~Tkach, A.~Kowdle, E.~Cooper, M.~Dou, S.~R. Fanello, G.~Fyffe, C.~Rhemann, J.~Taylor, P.~E. Debevec, and S.~Izadi, ``The relightables: Volumetric performance capture of humans with realistic relighting,'' \emph{ACM Transactions on Graphics}, vol.~38, no.~6, pp. 217:1--217:19, 2019.

\bibitem{Laffont2013}
P.-Y. Laffont, A.~Bousseau, and G.~Drettakis, ``Rich intrinsic image decomposition of outdoor scenes from multiple views,'' \emph{IEEE Transactions on Visualization and Computer Graphics (TVCG)}, vol.~19, no.~2, pp. 210--224, 2013.

\bibitem{Lensch2003}
H.~P. Lensch, J.~Lang, A.~M. Sa, and H.-P. Seidel, ``Planned sampling of spatially varying brdfs,'' \emph{Computer Graphics Forum}, 2003.

\bibitem{Park2020}
J.~J. Park, A.~Holynski, and S.~M. Seitz, ``Seeing the world in a bag of chips,'' in \emph{Proceedings of the IEEE Conference on Computer Vision and Pattern Recognition (CVPR)}, 2020.

\bibitem{Philip2019}
J.~Philip, M.~Gharbi, T.~Zhou, A.~A. Efros, and G.~Drettakis, ``Multi-view relighting using a geometry-aware network,'' \emph{ACM Transactions on Graphics}, vol.~38, no.~4, pp. 78:1--78:14, 2019.

\bibitem{Schmitt2020}
C.~Schmitt, S.~Donne, G.~Riegler, V.~Koltun, and A.~Geiger, ``On joint estimation of pose, geometry and svbrdf from a handheld scanner,'' in \emph{Proc. of the IEEE/CVF Conference on Computer Vision and Pattern Recognition (CVPR)}, 2020.

\bibitem{Zhang2021}
X.~Zhang, S.~R. Fanello, Y.-T. Tsai, T.~Sun, T.~Xue, R.~Pandey, S.~Orts-Escolano, P.~L. Davidson, C.~Rhemann, P.~E. Debevec, J.~T. Barron, R.~Ramamoorthi, and W.~T. Freeman, ``Neural light transport for relighting and view synthesis,'' \emph{ACM Transactions on Graphics}, vol.~40, no.~1, pp. 1--17, 2021.

\bibitem{mildenhall2020nerf}
B.~Mildenhall, P.~P. Srinivasan, M.~Tancik, J.~T. Barron, R.~Ramamoorthi, and R.~Ng, ``Nerf: Representing scenes as neural radiance fields for view synthesis,'' in \emph{ECCV}, 2020.

\bibitem{nerv2021}
P.~P. Srinivasan, B.~Deng, X.~Zhang, M.~Tancik, B.~Mildenhall, and J.~T. Barron, ``Nerv: Neural reflectance and visibility fields for relighting and view synthesis,'' in \emph{CVPR}, 2021.

\bibitem{boss2021nerd}
M.~Boss, R.~Braun, V.~Jampani, J.~T. Barron, C.~Liu, and H.~P. Lensch, ``Nerd: Neural reflectance decomposition from image collections,'' in \emph{IEEE International Conference on Computer Vision (ICCV)}, 2021.

\bibitem{knodt2021neural}
J.~Knodt, J.~Bartusek, S.-H. Baek, and F.~Heide, ``Neural ray-tracing: Learning surfaces and reflectance for relighting and view synthesis,'' 2021.

\bibitem{yang2022psnerf}
W.~Yang, G.~Chen, C.~Chen, Z.~Chen, and K.-Y.~K. Wong, ``Ps-nerf: Neural inverse rendering for multi-view photometric stereo,'' in \emph{European Conference on Computer Vision (ECCV)}, 2022.

\bibitem{boss2021neuralpil}
M.~Boss, V.~Jampani, R.~Braun, C.~Liu, J.~T. Barron, and H.~P. Lensch, ``Neural-pil: Neural pre-integrated lighting for reflectance decomposition,'' in \emph{Advances in Neural Information Processing Systems (NeurIPS)}, 2021.

\bibitem{physg2021}
K.~Zhang, F.~Luan, Q.~Wang, K.~Bala, and N.~Snavely, ``{PhySG}: {I}nverse rendering with spherical gaussians for physics-based material editing and relighting,'' in \emph{The IEEE/CVF Conference on Computer Vision and Pattern Recognition (CVPR)}, 2021.

\bibitem{zhang2021nerfactor}
X.~Zhang, P.~P. Srinivasan, B.~Deng, P.~Debevec, W.~T. Freeman, and J.~T. Barron, ``Nerfactor: Neural factorization of shape and reflectance under an unknown illumination,'' \emph{ACM Transactions on Graphics (ToG)}, vol.~40, no.~6, pp. 1--18, 2021.

\bibitem{Jin2023TensoIR}
H.~Jin, I.~Liu, P.~Xu, X.~Zhang, S.~Han, S.~Bi, X.~Zhou, Z.~Xu, and H.~Su, ``Tensoir: Tensorial inverse rendering,'' in \emph{Proceedings of the IEEE/CVF Conference on Computer Vision and Pattern Recognition (CVPR)}, 2023.

\bibitem{chen2022relighting}
Z.~Chen and Z.~Liu, ``Relighting4d: Neural relightable human from videos,'' in \emph{ECCV}, 2022.

\bibitem{lin2024relightable}
W.~Lin, C.~Zheng, J.-H. Yong, and F.~Xu, ``Relightable and animatable neural avatars from videos,'' in \emph{Proceedings of the AAAI Conference on Artificial Intelligence}, 2024, pp. 3486--3494.

\bibitem{sun2023neural}
W.~Sun, Y.~Che, H.~Huang, and Y.~Guo, ``Neural reconstruction of relightable human model from monocular video,'' in \emph{Proceedings of the IEEE/CVF International Conference on Computer Vision}, 2023, pp. 397--407.

\bibitem{wang2021neus}
P.~Wang, L.~Liu, Y.~Liu, C.~Theobalt, T.~Komura, and W.~Wang, ``Neus: Learning neural implicit surfaces by volume rendering for multi-view reconstruction,'' \emph{arXiv preprint arXiv:2106.10689}, 2021.

\bibitem{wang2024intrinsicavatar}
S.~Wang, B.~Anti{\'c}, A.~Geiger, and S.~Tang, ``Intrinsicavatar: Physically based inverse rendering of dynamic humans from monocular videos via explicit ray tracing,'' in \emph{Proceedings of the IEEE/CVF Conference on Computer Vision and Pattern Recognition (CVPR)}.\hskip 1em plus 0.5em minus 0.4em\relax IEEE, 2024.

\bibitem{xu2024relightable}
Z.~Xu, S.~Peng, C.~Geng, L.~Mou, Z.~Yan, J.~Sun, H.~Bao, and X.~Zhou, ``Relightable and animatable neural avatar from sparse-view video,'' in \emph{CVPR}, 2024.

\bibitem{mueller2022instant}
\BIBentryALTinterwordspacing
T.~M\"uller, A.~Evans, C.~Schied, and A.~Keller, ``Instant neural graphics primitives with a multiresolution hash encoding,'' \emph{ACM Trans. Graph.}, vol.~41, no.~4, pp. 102:1--102:15, Jul. 2022. [Online]. Available: \url{https://doi.org/10.1145/3528223.3530127}
\BIBentrySTDinterwordspacing

\bibitem{kerbl3Dgaussians}
\BIBentryALTinterwordspacing
B.~Kerbl, G.~Kopanas, T.~Leimk{\"u}hler, and G.~Drettakis, ``3d gaussian splatting for real-time radiance field rendering,'' \emph{ACM Transactions on Graphics}, vol.~42, no.~4, July 2023. [Online]. Available: \url{https://repo-sam.inria.fr/fungraph/3d-gaussian-splatting/}
\BIBentrySTDinterwordspacing

\bibitem{Huang2DGS2024}
B.~Huang, Z.~Yu, A.~Chen, A.~Geiger, and S.~Gao, ``2d gaussian splatting for geometrically accurate radiance fields,'' \emph{SIGGRAPH}, 2024.

\bibitem{instant_nvr}
C.~Geng, S.~Peng, Z.~Xu, H.~Bao, and X.~Zhou, ``Learning neural volumetric representations of dynamic humans in minutes,'' in \emph{CVPR}, 2023.

\bibitem{jiang2022instantavatar}
T.~Jiang, X.~Chen, J.~Song, and O.~Hilliges, ``Instantavatar: Learning avatars from monocular video in 60 seconds,'' \emph{arXiv}, 2022.

\bibitem{Chen2023PAMI}
X.~Chen, T.~Jiang, J.~Song, M.~Rietmann, A.~Geiger, M.~J. Black, and O.~Hilliges, ``Fast-snarf: A fast deformer for articulated neural fields,'' \emph{Pattern Analysis and Machine Intelligence (PAMI)}, 2023.

\bibitem{chen2021snarf}
X.~Chen, Y.~Zheng, M.~J. Black, O.~Hilliges, and A.~Geiger, ``Snarf: Differentiable forward skinning for animating non-rigid neural implicit shapes,'' in \emph{International Conference on Computer Vision (ICCV)}, 2021.

\bibitem{SMPL}
M.~Loper, N.~Mahmood, J.~Romero, G.~Pons-Moll, and M.~J. Black, ``{SMPL}: A skinned multi-person linear model,'' \emph{ACM Trans. Graphics (Proc. SIGGRAPH Asia)}, vol.~34, no.~6, pp. 248:1--248:16, Oct. 2015.

\bibitem{Lei2024GART}
J.~Lei, Y.~Wang, G.~Pavlakos, L.~Liu, and K.~Daniilidis, ``Gart: Gaussian articulated template models,'' in \emph{Proceedings of the IEEE/CVF Conference on Computer Vision and Pattern Recognition (CVPR)}, 2024.

\bibitem{li2024animatablegaussians}
Z.~Li, Z.~Zheng, L.~Wang, and Y.~Liu, ``Animatable gaussians: Learning pose-dependent gaussian maps for high-fidelity human avatar modeling,'' in \emph{Proceedings of the IEEE/CVF Conference on Computer Vision and Pattern Recognition (CVPR)}, 2024.

\bibitem{zheng2024gpsgaussian}
S.~Zheng, B.~Zhou, R.~Shao, B.~Liu, S.~Zhang, L.~Nie, and Y.~Liu, ``Gps-gaussian: Generalizable pixel-wise 3d gaussian splatting for real-time human novel view synthesis,'' in \emph{Proceedings of the IEEE/CVF Conference on Computer Vision and Pattern Recognition (CVPR)}, 2024.

\bibitem{GauHuman}
S.~Hu and Z.~Liu, ``Gauhuman: Articulated gaussian splatting for real-time 3d human rendering,'' \emph{arXiv preprint}, 2023.

\bibitem{qian20233dgsavatar}
Z.~Qian, S.~Wang, M.~Mihajlovic, A.~Geiger, and S.~Tang, ``3dgs-avatar: Animatable avatars via deformable 3d gaussian splatting,'' in \emph{CVPR}, 2024.

\bibitem{yuan2023gavatar}
Y.~Yuan, X.~Li, Y.~Huang, S.~De~Mello, K.~Nagano, J.~Kautz, and U.~Iqbal, ``Gavatar: Animatable 3d gaussian avatars with implicit mesh learning,'' \emph{arXiv preprint arXiv:2312.11461}, 2023.

\bibitem{zhu2023ash}
H.~Pang, H.~Zhu, A.~Kortylewski, C.~Theobalt, and M.~Habermann, ``Ash: Animatable gaussian splats for efficient and photoreal human rendering,'' \emph{arXiv preprint arXiv:2312.05941}, 2023.

\bibitem{liang2023gs}
Z.~Liang, Q.~Zhang, Y.~Feng, Y.~Shan, and K.~Jia, ``Gs-ir: 3d gaussian splatting for inverse rendering,'' \emph{arXiv preprint arXiv:2311.16473}, 2023.

\bibitem{shi2023gir}
Y.~Shi, Y.~Wu, C.~Wu, X.~Liu, C.~Zhao, H.~Feng, J.~Liu, L.~Zhang, J.~Zhang, B.~Zhou, E.~Ding, and J.~Wang, ``Gir: 3d gaussian inverse rendering for relightable scene factorization,'' \emph{Arxiv}, 2023.

\bibitem{jiang2023gaussianshader}
Y.~Jiang, J.~Tu, Y.~Liu, X.~Gao, X.~Long, W.~Wang, and Y.~Ma, ``Gaussianshader: 3d gaussian splatting with shading functions for reflective surfaces,'' \emph{arXiv preprint arXiv:2311.17977}, 2023.

\bibitem{R3DG2023}
J.~Gao, C.~Gu, Y.~Lin, H.~Zhu, X.~Cao, L.~Zhang, and Y.~Yao, ``Relightable 3d gaussian: Real-time point cloud relighting with brdf decomposition and ray tracing,'' \emph{arXiv:2311.16043}, 2023.

\bibitem{Bolanos2024GaussianShadowCasting}
L.~Bolanos, S.-Y. Su, and H.~Rhodin, ``Gaussian shadow casting for neural characters,'' in \emph{Proceedings of the IEEE/CVF Conference on Computer Vision and Pattern Recognition (CVPR)}, 2024.

\bibitem{libigl-python-bindings-chapter5}
``Chapter 5 - igl,'' \url{https://libigl.github.io/libigl-python-bindings/tut-chapter5/}, 2023.

\bibitem{zwicker2004perspective}
M.~Zwicker, J.~Rasanen, M.~Botsch, C.~Dachsbacher, and M.~Pauly, ``Perspective accurate splatting,'' in \emph{Proceedings-Graphics Interface}, 2004, pp. 247--254.

\bibitem{Bi2020}
S.~Bi, Z.~Xu, P.~Srinivasan, B.~Mildenhall, K.~Sunkavalli, M.~Ha{\v s}an, Y.~Hold-Geoffroy, D.~Kriegman, and R.~Ramamoorthi, ``Neural reflectance fields for appearance acquisition,'' \emph{arXiv}, vol. 2008.03824, 2020.

\bibitem{Boss2020}
M.~Boss, V.~Jampani, K.~Kim, H.~Lensch, and J.~Kautz, ``Two-shot spatially-varying brdf and shape estimation,'' in \emph{Proc. of the IEEE/CVF Conference on Computer Vision and Pattern Recognition (CVPR)}, 2020, pp. 3982--3991.

\bibitem{Gardner2003}
A.~Gardner, C.~Tchou, T.~Hawkins, and P.~Debevec, ``Linear light source reflectometry,'' \emph{ACM Transactions on Graphics}, vol.~22, no.~3, pp. 749--758, 2003.

\bibitem{Ghosh2009}
A.~Ghosh, T.~Chen, P.~Peers, C.~A. Wilson, and P.~Debevec, ``Estimating specular roughness and anisotropy from second order spherical gradient illumination,'' \emph{Computer Graphics Forum}, vol.~28, no.~4, pp. 1161--1170, 2009.

\bibitem{Guarnera2016}
D.~Guarnera, G.~C. Guarnera, A.~Ghosh, C.~Denk, and M.~Glencross, ``Brdf representation and acquisition,'' in \emph{Proceedings of the 37th Annual Conference of the European Association for Computer Graphics: State of the Art Reports}, 2016, pp. 625--650.

\bibitem{Haindl2013}
M.~Haindl and J.~Filip, \emph{Visual Texture}.\hskip 1em plus 0.5em minus 0.4em\relax Springer-Verlag, 2013.

\bibitem{Weinmann2015}
M.~Weinmann and R.~Klein, ``Advances in geometry and reflectance acquisition (course notes),'' in \emph{SIGGRAPH Asia Courses}, 2015.

\bibitem{lyu2022neural}
L.~Lyu, A.~Tewari, T.~Leimkuehler, M.~Habermann, and C.~Theobalt, ``Neural radiance transfer fields for relightable novel-view synthesis with global illumination,'' \emph{arXiv preprint arXiv:2207.10662}, 2022.

\bibitem{lyu2023diffusion}
L.~Lyu, A.~Tewari, M.~Habermann, S.~Saito, M.~Zollh{"o}fer, T.~Leimk{"u}hler, and C.~Theobalt, ``Diffusion posterior illumination for ambiguity-aware inverse rendering,'' \emph{ACM Transactions on Graphics (TOG)}, vol.~42, no.~2, pp. 1--14, 2023.

\bibitem{yao2022neilf}
Y.~Yao, J.~Zhang, J.~Liu, Y.~Qu, T.~Fang, D.~McKinnon, Y.~Tsin, and L.~Quan, ``Neilf: Neural incident light field for physically-based material estimation,'' in \emph{European Conference on Computer Vision (ECCV)}, 2022.

\bibitem{boss2022samurai}
M.~Boss, A.~Engelhardt, A.~Kar, Y.~Li, D.~Sun, J.~T. Barron, H.~P. Lensch, and V.~Jampani, ``{SAMURAI}: {S}hape {A}nd {M}aterial from {U}nconstrained {R}eal-world {A}rbitrary {I}mage collections,'' in \emph{Advances in Neural Information Processing Systems (NeurIPS)}, 2022.

\bibitem{Zhang_2022_CVPR}
Y.~Zhang, J.~Sun, X.~He, H.~Fu, R.~Jia, and X.~Zhou, ``Modeling indirect illumination for inverse rendering,'' in \emph{Proceedings of the IEEE/CVF Conference on Computer Vision and Pattern Recognition (CVPR)}, June 2022, pp. 18\,643--18\,652.

\bibitem{liu2022neuray}
Y.~Liu, S.~Peng, L.~Liu, Q.~Wang, P.~Wang, T.~Christian, X.~Zhou, and W.~Wang, ``Neural rays for occlusion-aware image-based rendering,'' in \emph{CVPR}, 2022.

\bibitem{Zhang2023}
Y.~Zhang, T.~Xu, J.~Yu, Y.~Ye, J.~Wang, Y.~Jing, and W.~Yang, ``Nemf: Inverse volume rendering with neural microflake field,'' in \emph{Proceedings of the IEEE International Conference on Computer Vision (ICCV)}, 2023.

\bibitem{sitzmann2019siren}
V.~Sitzmann, J.~N. Martel, A.~W. Bergman, D.~B. Lindell, and G.~Wetzstein, ``Implicit neural representations with periodic activation functions,'' in \emph{Proc. NeurIPS}, 2020.

\bibitem{Chen2022ECCV}
A.~Chen, Z.~Xu, A.~Geiger, and H.~Su, ``Tensorf: Tensorial radiance fields,'' in \emph{European Conference on Computer Vision (ECCV)}, 2022.

\bibitem{liu2023nero}
Y.~Liu, P.~Wang, C.~Lin, X.~Long, J.~Wang, L.~Liu, T.~Komura, and W.~Wang, ``Nero: Neural geometry and brdf reconstruction of reflective objects from multiview images,'' in \emph{TOG}, 2023.

\bibitem{mao2023neuspir}
S.~Mao, C.~Wu, Z.~Shen, and L.~Zhang, ``Neus-pir: Learning relightable neural surface using pre-integrated rendering,'' \emph{arXiv preprint arXiv:2306.07632}, 2023.

\bibitem{iron_2022}
K.~Zhang, F.~Luan, Z.~Li, and N.~Snavely, ``Iron: Inverse rendering by optimizing neural sdfs and materials from photometric images,'' in \emph{IEEE Conf. Comput. Vis. Pattern Recog.}, 2022.

\bibitem{yang2023sireir}
Z.~Yang, Y.~Chen, X.~Gao, Y.~Yuan, Y.~Wu, X.~Zhou, and X.~Jin, ``Sire-ir: Inverse rendering for brdf reconstruction with shadow and illumination removal in high-illuminance scenes,'' \emph{arXiv preprint arXiv:2310.13030}, 2023.

\bibitem{wang2024nep}
H.~Wang, W.~Hu, L.~Zhu, and R.~Lau, ``Inverse rendering of glossy objects via the neural plenoptic function and radiance fields,'' in \emph{CVPR}, 2024.

\bibitem{shen2021dmtet}
T.~Shen, J.~Gao, K.~Yin, M.-Y. Liu, and S.~Fidler, ``Deep marching tetrahedra: a hybrid representation for high-resolution 3d shape synthesis,'' in \emph{Advances in Neural Information Processing Systems (NeurIPS)}, 2021.

\bibitem{Munkberg_2022_CVPR}
J.~Munkberg, J.~Hasselgren, T.~Shen, J.~Gao, W.~Chen, A.~Evans, T.~M\"uller, and S.~Fidler, ``{Extracting Triangular 3D Models, Materials, and Lighting From Images},'' in \emph{Proceedings of the IEEE/CVF Conference on Computer Vision and Pattern Recognition (CVPR)}, June 2022, pp. 8280--8290.

\bibitem{hasselgren2022nvdiffrecmc}
J.~Hasselgren, N.~Hofmann, and J.~Munkberg, ``{Shape, Light, and Material Decomposition from Images using Monte Carlo Rendering and Denoising},'' \emph{arXiv:2206.03380}, 2022.

\bibitem{Yu2024GOF}
Z.~Yu, T.~Sattler, and A.~Geiger, ``Gaussian opacity fields: Efficient high-quality compact surface reconstruction in unbounded scenes,'' \emph{arXiv preprint arXiv:2404.10772}, 2024.

\bibitem{Dai2024GaussianSurfels}
P.~Dai, J.~Xu, W.~Xie, X.~Liu, H.~Wang, and W.~Xu, ``High-quality surface reconstruction using gaussian surfels,'' in \emph{SIGGRAPH 2024 Conference Papers}.\hskip 1em plus 0.5em minus 0.4em\relax Association for Computing Machinery, 2024.

\bibitem{peng2021animatable}
S.~Peng, J.~Dong, Q.~Wang, S.~Zhang, Q.~Shuai, X.~Zhou, and H.~Bao, ``Animatable neural radiance fields for modeling dynamic human bodies,'' in \emph{Proceedings of the IEEE/CVF International Conference on Computer Vision}, 2021, pp. 14\,314--14\,323.

\bibitem{peng2021neural}
S.~Peng, Y.~Zhang, Y.~Xu, Q.~Wang, Q.~Shuai, H.~Bao, and X.~Zhou, ``Neural body: Implicit neural representations with structured latent codes for novel view synthesis of dynamic humans,'' in \emph{CVPR}, 2021.

\bibitem{liu2021neuralactor}
L.~Liu, M.~Habermann, V.~Rudnev, K.~Sarkar, J.~Gu, and C.~Theobalt, ``Neural actor: Neural free-view synthesis of human actors with pose control,'' \emph{ACM SIGGRAPH Asia}, 2021.

\bibitem{zhi2022dual}
Y.~Zhi, S.~Qian, X.~Yan, and S.~Gao, ``Dual-space nerf: Learning animatable avatars and scene lighting in separate spaces,'' in \emph{International Conference on 3D Vision (3DV)}, Sep. 2022.

\bibitem{weng_humannerf_2022_cvpr}
C.-Y. Weng, B.~Curless, P.~P. Srinivasan, J.~T. Barron, and I.~Kemelmacher-Shlizerman, ``Human{N}e{RF}: Free-viewpoint rendering of moving people from monocular video,'' in \emph{Proceedings of the IEEE/CVF Conference on Computer Vision and Pattern Recognition (CVPR)}, June 2022, pp. 16\,210--16\,220.

\bibitem{huang2024tech}
Y.~Huang, H.~Yi, Y.~Xiu, T.~Liao, J.~Tang, D.~Cai, and J.~Thies, ``{TeCH: Text-guided Reconstruction of Lifelike Clothed Humans},'' in \emph{International Conference on 3D Vision (3DV)}, 2024.

\bibitem{huang2022elicit}
Y.~Huang, H.~Yi, W.~Liu, H.~Wang, B.~Wu, W.~Wang, B.~Lin, D.~Zhang, and D.~Cai, ``One-shot implicit animatable avatars with model-based priors,'' in \emph{IEEE Conference on Computer Vision (ICCV)}, 2023.

\bibitem{ARAH_eccv_2022}
S.~Wang, K.~Schwarz, A.~Geiger, and S.~Tang, ``Arah: Animatable volume rendering of articulated human sdfs,'' in \emph{European Conference on Computer Vision}, 2022.

\bibitem{debsdf2024}
Y.~Xiao, J.~Xu, Z.~Yu, and S.~Gao, ``Debsdf: Delving into the details and bias of neural indoor scene reconstruction,'' \emph{IEEE Transactions on Pattern Analysis and Machine Intelligence (TPAMI)}, 2024.

\bibitem{wang2023neus2}
Y.~Wang, Q.~Han, M.~Habermann, K.~Daniilidis, C.~Theobalt, and L.~Liu, ``Neus2: Fast learning of neural implicit surfaces for multi-view reconstruction,'' in \emph{Proceedings of the IEEE/CVF International Conference on Computer Vision (ICCV)}.\hskip 1em plus 0.5em minus 0.4em\relax IEEE, 2023.

\bibitem{hu2024gaussianavatar}
L.~Hu, H.~Zhang, Y.~Zhang, B.~Zhou, B.~Liu, S.~Zhang, and L.~Nie, ``Gaussianavatar: Towards realistic human avatar modeling from a single video via animatable 3d gaussians,'' in \emph{IEEE/CVF Conference on Computer Vision and Pattern Recognition (CVPR)}, 2024.

\bibitem{wen2024gomavatar}
J.~Wen, X.~Zhao, Z.~Ren, A.~Schwing, and S.~Wang, ``{GoMAvatar: Efficient Animatable Human Modeling from Monocular Video Using Gaussians-on-Mesh},'' in \emph{CVPR}, 2024.

\bibitem{zhi2025strugauavatar}
Y.~Zhi, W.~Sun, J.~Chang, C.~Ye, W.~Feng, and X.~Han, ``{StruGauAvatar: Learning Structured 3D Gaussians for Animatable Avatars from Monocular Videos},'' \emph{IEEE Transactions on Visualization and Computer Graphics}, 2025.

\bibitem{kim2024switchlight}
H.~Kim, M.~Jang, W.~Yoon, J.~Lee, D.~Na, and S.~Woo, ``Switchlight: Co-design of physics-driven architecture and pre-training framework for human portrait relighting,'' \emph{arXiv preprint arXiv:2024.xxxxx}, 2024.

\bibitem{pandey2021total}
R.~Pandey, S.~Orts~Escolano, C.~Legendre, C.~Haene, S.~Bouaziz, C.~Rhemann, P.~Debevec, and S.~Fanello, ``Total relighting: learning to relight portraits for background replacement,'' \emph{ACM Transactions on Graphics (TOG)}, vol.~40, no.~4, pp. 1--21, 2021.

\bibitem{sun2019single}
T.~Sun, J.~T. Barron, Y.-T. Tsai, Z.~Xu, X.~Yu, G.~Fyffe, C.~Rhemann, J.~Busch, P.~Debevec, and R.~Ramamoorthi, ``Single image portrait relighting,'' \emph{ACM Transactions on Graphics (TOG)}, vol.~38, no.~4, pp. 1--12, 2019.

\bibitem{yeh2022learning}
Y.-Y. Yeh, K.~Nagano, S.~Khamis, J.~Kautz, M.-Y. Liu, and T.-C. Wang, ``Learning to relight portrait images via a virtual light stage and synthetic-to-real adaptation,'' \emph{ACM Transactions on Graphics (TOG)}, vol.~41, no.~6, pp. 1--21, 2022.

\bibitem{saito2024rgca}
S.~Saito, G.~Schwartz, T.~Simon, J.~Li, and G.~Nam, ``Relightable gaussian codec avatars,'' in \emph{CVPR}, 2024.

\bibitem{iqbal2023rana}
U.~Iqbal, A.~Caliskan, K.~Nagano, S.~Khamis, P.~Molchanov, and J.~Kautz, ``Rana: Relightable articulated neural avatars,'' in \emph{ICCV}, 2023.

\bibitem{relightneuralactor2024}
D.~Luvizon, V.~Golyanik, A.~Kortylewski, M.~Habermann, and C.~Theobalt, ``Relightable neural actor with intrinsic decomposition and pose control,'' \emph{arXiv}, 2023.

\bibitem{qian2022unif}
S.~Qian, J.~Xu, Z.~Liu, L.~Ma, and S.~Gao, ``Unif: United neural implicit functions for clothed human reconstruction and animation,'' in \emph{European Conference on Computer Vision}.\hskip 1em plus 0.5em minus 0.4em\relax Springer, 2022, pp. 121--137.

\bibitem{ClothCap}
\BIBentryALTinterwordspacing
G.~Pons-Moll, S.~Pujades, S.~Hu, and M.~Black, ``Clothcap: Seamless 4d clothing capture and retargeting,'' \emph{ACM Transactions on Graphics, (Proc. SIGGRAPH)}, vol.~36, no.~4, 2017. [Online]. Available: \url{http://dx.doi.org/10.1145/3072959.3073711}
\BIBentrySTDinterwordspacing

\bibitem{AMASS_ICCV_2019}
N.~Mahmood, N.~Ghorbani, N.~F. Troje, G.~Pons-Moll, and M.~J. Black, ``{AMASS}: Archive of motion capture as surface shapes,'' in \emph{International Conference on Computer Vision}, Oct. 2019, pp. 5442--5451.

\bibitem{li2024animatable}
Z.~Li, Y.~Sun, Z.~Zheng, L.~Wang, S.~Zhang, and Y.~Liu, ``Animatable and relightable gaussians for high-fidelity human avatar modeling,'' \emph{arXiv preprint arXiv:2311.16096v4}, 2024.

\bibitem{Karras2019stylegan2}
T.~Karras, S.~Laine, M.~Aittala, J.~Hellsten, J.~Lehtinen, and T.~Aila, ``Analyzing and improving the image quality of {StyleGAN},'' in \emph{Proc. CVPR}, 2020.

\bibitem{PhysAvatar2024}
Y.~Zheng, Q.~Zhao, G.~Yang, W.~Yifan, D.~Xiang, F.~Dubost, D.~Lagun, T.~Beeler, F.~Tombari, L.~Guibas, and G.~Wetzstein, ``Physavatar: Learning the physics of dressed 3d avatars from visual observations,'' \emph{arXiv preprint arXiv:2404.04421}, 2024.

\bibitem{hu2023gauhuman}
S.~Hu and Z.~Liu, ``Gauhuman: Articulated gaussian splatting from monocular human videos,'' \emph{arXiv preprint arXiv:}, 2023.

\bibitem{Karis2013RealShading}
B.~Karis and E.~Games, ``Real shading in unreal engine 4,'' in \emph{Proceedings of Physically Based Shading Theory Practice}, vol.~4, no.~3, 2013, p.~1.

\bibitem{Burley2012PhysicallyBasedShading}
B.~Burley and W.~D.~A. Studios, ``Physically-based shading at disney,'' in \emph{ACM SIGGRAPH}, vol. 2012, 2012, pp. 1--7.

\bibitem{GGX}
B.~Walter, S.~R. Marschner, H.~Li, and K.~E. Torrance, ``Microfacet models for refraction through rough surfaces,'' in \emph{Proceedings of the 18th Eurographics Conference on Rendering Techniques}, ser. EGSR'07.\hskip 1em plus 0.5em minus 0.4em\relax Goslar, DEU: Eurographics Association, 2007, p. 195–206.

\bibitem{peoplesnapshot}
T.~Alldieck, M.~Magnor, W.~Xu, C.~Theobalt, and G.~Pons-Moll, ``Video based reconstruction of 3d people models,'' in \emph{{IEEE}/{CVF} Conference on Computer Vision and Pattern Recognition ({CVPR})}, Jun 2018, pp. 8387--8397.

\bibitem{fang2021mirrored}
Q.~Fang, Q.~Shuai, J.~Dong, H.~Bao, and X.~Zhou, ``Reconstructing 3d human pose by watching humans in the mirror,'' in \emph{CVPR}, 2021.

\bibitem{polyhaven_hdris}
\BIBentryALTinterwordspacing
{Poly Haven}, ``Hdris,'' 2024, accessed: 2024-06-13. [Online]. Available: \url{https://polyhaven.com/hdris}
\BIBentrySTDinterwordspacing

\bibitem{li2021learn}
R.~Li, S.~Yang, D.~A. Ross, and A.~Kanazawa, ``Learn to dance with aist++: Music conditioned 3d dance generation,'' 2021.

\bibitem{CAPE_CVPR}
Q.~Ma, J.~Yang, A.~Ranjan, S.~Pujades, G.~Pons-Moll, S.~Tang, and M.~J. Black, ``{Learning to Dress 3D People in Generative Clothing},'' in \emph{Computer Vision and Pattern Recognition (CVPR)}, June 2020.

\bibitem{zhou2018open3d}
\BIBentryALTinterwordspacing
Q.-Y. Zhou, J.~Park, and V.~Koltun, ``Open3d: A modern library for 3d data processing,'' 2018, cite arxiv:1801.09847Comment: http://www.open3d.org. [Online]. Available: \url{http://arxiv.org/abs/1801.09847}
\BIBentrySTDinterwordspacing

\bibitem{Xu2024}
Y.~Xu, P.~Chandran, S.~Weiss, M.~Gross, G.~Zoss, and D.~Bradley, ``Artist-friendly relightable and animatable neural heads,'' in \emph{Proceedings of the IEEE/CVF Conference on Computer Vision and Pattern Recognition (CVPR)}, 2024.

\bibitem{qian2023gaussianavatars}
S.~Qian, T.~Kirschstein, L.~Schoneveld, D.~Davoli, S.~Giebenhain, and M.~Nie\ss{}ner, ``Gaussianavatars: Photorealistic head avatars with rigged 3d gaussians,'' \emph{arXiv preprint arXiv:2312.02069}, 2023.

\bibitem{saito2024relightable}
S.~Saito, G.~Schwartz, T.~Simon, J.~Li, and G.~Nam, ``Relightable gaussian codec avatars,'' in \emph{Proceedings of the IEEE/CVF Conference on Computer Vision and Pattern Recognition (CVPR)}, 2024.

\bibitem{cai2023smplerx}
Z.~Cai, W.~Yin, A.~Zeng, C.~Wei, Q.~Sun, W.~Yanjun, H.~E. Pang, H.~Mei, M.~Zhang, L.~Zhang, C.~C. Loy, L.~Yang, and Z.~Liu, ``{SMPLer-X}: Scaling up expressive human pose and shape estimation,'' in \emph{Advances in Neural Information Processing Systems}, 2023.

\bibitem{yin2025smplest}
W.~Yin, Z.~Cai, R.~Wang, A.~Zeng, C.~Wei, Q.~Sun, H.~Mei, Y.~Wang, H.~E. Pang, M.~Zhang, L.~Zhang, C.~C. Loy, A.~Yamashita, L.~Yang, and Z.~Liu, ``{SMPLest-X}: Ultimate scaling for expressive human pose and shape estimation,'' \emph{arXiv preprint arXiv:2501.09782}, 2025.

\bibitem{careagaColorful}
C.~Careaga and Y.~Aksoy, ``Colorful diffuse intrinsic image decomposition in the wild,'' \emph{ACM Trans. Graph.}, vol.~43, no.~6, 2024.

\bibitem{careagaIntrinsic}
------, ``Intrinsic image decomposition via ordinal shading,'' \emph{ACM Trans. Graph.}, vol.~43, no.~1, 2023.

\bibitem{chen2024intrinsicanything}
C.~Xi, P.~Sida, Y.~Dongchen, L.~Yuan, P.~Bowen, L.~Chengfei, and Z.~Xiaowei, ``Intrinsicanything: Learning diffusion priors for inverse rendering under unknown illumination,'' \emph{arxiv: 2404.11593}, 2024.

\end{thebibliography}
